%% file: main.tex
\documentclass{article}
\usepackage[T1]{fontenc}    
\usepackage[utf8]{inputenc} 
\usepackage{amsfonts}       
\usepackage{arxiv}
\usepackage{booktabs}       
\usepackage{doi}
\usepackage{float}
\usepackage{graphicx}
\usepackage{hyperref}       
\usepackage{microtype}      
\usepackage[sort,comma,authoryear,round]{natbib}
\usepackage{nicefrac}       
\usepackage{subcaption}
\usepackage{url}            

\title{SideSeeing: A multimodal dataset and collection of tools for sidewalk assessment}

\author{
R. J. P. Damaceno\\
    Department of Computer Science\\
    Institute of Mathematics and Statistics\\
    University of São Paulo (USP)
\And
L. Ferreira\\
    Department of Computer Science\\
    University of Illinois Chicago (UIC)\\
\And
F. Miranda\\
    Department of Computer Science\\
    University of Illinois Chicago (UIC)\\
\And
M. Hosseini\\
    City Form Lab\\
    Massachusetts Institute of Technology (MIT)
\And
R. M. Cesar-Jr.\\
    Department of Computer Science\\
    Institute of Mathematics and Statistics\\
    University of São Paulo (USP)
}

\renewcommand{\shorttitle}{\textit{SideSeeing}}

\hypersetup{
pdftitle={A multimodal dataset and collection of tools for sidewalk assessment},
pdfsubject={cs.CV, cs.CY},
pdfauthor={R. J. P. Damaceno, L. Ferreira, F. Miranda, M. Hosseini, R. M. Cesar-Jr.},
pdfkeywords={SideSeeing, Sidewalk, Dataset},
}

\begin{document}

\maketitle

\begin{abstract}
This paper introduces SideSeeing, a novel initiative that provides tools and datasets for assessing the built environment. We present a framework for street-level data acquisition, loading, and analysis. Using the framework, we collected a novel dataset that integrates synchronized video footaged captured from chest-mounted mobile devices with sensor data (accelerometer, gyroscope, magnetometer, and GPS). Each data sample represents a path traversed by a user filming sidewalks near hospitals in Brazil and the USA. The dataset encompasses three hours of content covering 12 kilometers around nine hospitals, and includes 325,000 video frames with corresponding sensor data. Additionally, we present a novel 68-element taxonomy specifically created for sidewalk scene identification. SideSeeing is a step towards a suite of tools that urban experts can use to perform in-depth sidewalk accessibility evaluations. SideSeeing data and tools are publicly available at \url{https://sites.usp.br/sideseeing/}.
\end{abstract}

\section{Introduction}

In recent years, urban informatics and urban computing have opened new horizons for tackling a number societal problems integral to urban planning and design~\citep{biljecki2021sviandgis, wang2022geoai, marasinghe2023computer, miranda2024sota3d} such as urban accessibility~\citep{saha2019project}, risk assessment and mapping~\citep{darabi2019urban}, climate change~\citep{cowls2023ai}, and heat exposure~\citep{hsu2021disproportionate}. 

The convergence of digital technologies with urban infrastructure has become essential in this area. Data collected from sensors installed in city infrastructure or from individuals' devices facilitate the creation of models for informed decision-making~\citep{kontokosta2021urban, shi2021urban}. These models can characterize various aspects of urban life, including sidewalk surfaces~\citep{hosseini2022citysurfaces}, vegetation presence~\citep{biljecki2023greenery}, region-specific noise levels~\citep{rulff2022urbanrhapsody, zhao2023soundscapes}, and land use~\citep{miranda2020urbanmosaic}.
%
These data can be represented in various formats, for instance, visual data captured by video cameras, sound data captured by microphones, textual data extracted from region-related databases, and temporal data obtained from sensors in mobile devices. Real-world urban informatics scenarios are inherently multimodal, integrating these diverse data types to provide comprehensive insights.

Despite the recent surge in urban-centric research, there remains a gap in the availability of comprehensive datasets describing urban public spaces, particularly those dedicated to pedestrians~\citep{deitz2021squeaky, hosseini2023mapping}.
While some datasets may offer information about sidewalks accessibility~\citep{saha2019project, park2020sideguide}, pedestrian networks~\citep{hosseini2023mapping}, sidewalk surface materials~\citep{hosseini2022citysurfaces},there is a lack of fine-grained, ground-level data on sidewalk conditions that reflect the lived experiences of their users. 
Integrating multimodal data has the potential to improve our understanding of accessibility in urban areas.

To address this gap, we introduce SideSeeing, a novel initiative that provides tools and datasets for assessing the built environment.
SideSeeing includes a framework for street-level data acquisition, loading, and analysis.
It also includes a novel dataset that integrates synchronized video footaged captured from chest-mounted mobile devices with sensor data (accelerometer, gyroscope, magnetometer, and GPS).
The dataset is focused on environments near hospitals, areas that are of particularly importance for public health and accessibility.
As \cite{seetharaman2024influence} note, gaining a better understanding of the built environment, particularly outdoors spaces, can provide key information to the design of accessible public spaces.
Our dataset provides detailed information about surface types and conditions.
By collecting synchronized data from multiple sensors and sources, including video, GPS, and IMU sensors available on mobile devices, we offer a multimodal dataset for analyzing urban environments and assessing accessibility in different contexts.


The main contributions of this paper are: 1) A multimodal dataset composed of urban scenes representing sidewalks near hospitals, called SideSeeing Hospital Dataset; 2) An open-source Android application for collecting synchronized multimodal data using smartphones; 3) A novel taxonomy specifically designed for characterize urban scenes focused on sidewalks; and 4) A Python library for loading and analyzing datasets created with our framework.

This paper is organized as follows: In Section~\ref{sec:related}, we briefly review related work.
In Section~\ref{sec:methods}, we introduce our methodology, including equipment, collection protocol, and data specification.
In Section~\ref{sec:dataset}, we present the SideSeeing Hospital Dataset.
In Section~\ref{sec:tools}, we present our open-source tools.
In Section~\ref{sec:conclusions}, we present our conclusions and future work.

\section{Related Work}
\label{sec:related}
Here, we provide a review of works that leverage computer vision to assess the built environment, particularly focusing on sidewalks.
When describing scenes of sidewalks, a key research topic is related to navigation assistance. In this regard, the work by \cite{kuriakose2023deepnavi} developed a mobile application to assist visually impaired people in identifying the built environment. Their solution, worn as a vest with a built-in smartphone, captures video of the environment. Computer vision modules then process the captured frames, and results are conveyed to the user through audio feedback. The proposed solution is built on the EfficientDet family model, does not require an internet connection, and can detect 20 different types of obstacles present in the scenes, providing both their size and distance to the user. Additionally, it can classify scenes into 20 categories relevant to the assistance navigation domain. 
%

\cite{choi2022integrated} employed a solution based on the YOLO family to assess the degree of walkway breakage. The authors defined criteria to categorize the breakage severity as ``very bad'', ``bad'', ``normal'', and ``good''. They found that the model achieved a 92\% accuracy in detecting walkway breakages. This approach has the potential to significantly reduce the time required for walkway assessments.

Another interesting point discussed in several papers is the analysis of the built environment in relation to the implementation of urban structures that promote social inclusion. As shown in \cite{shashiki2023walkability}, even when an urban structure is improved, the surrounding environment may not necessarily benefit from the solution. This is exemplified by the study's examination of a Brazilian infrastructure project called CEU, a governmental initiative aimed at promoting social inclusion. The work found that the walkability of the surrounding environment (e.g., sidewalks) did not improved alongside the other improvements.

With the perspective of automatize evaluation of the cities, the study by \cite{zhang2023oasis} developed a wheelchair-mounted prototype called OASIS that captures ground-level data to evaluate sidewalks. This system utilizes a combination of hardware, including a stereo camera, three-axis camera jig, GPS receiver, and a computation device. OASIS can segment infrastructure and street furniture, generating data for sidewalk assessment.
With a similar objective, but leveraging low-cost sensors, \cite{ng2023deep} studied the ability to detect irregular walking surfaces using only accelerometer data. Their model, based on a LSTM network, utilizes gait data gathered from a single wearable accelerometer to automatically identify these surfaces (e.g., well-paved, grassed, obstructed with objects, uneven, and covered with debris). The authors reported an average AUC of 88\% for classification performance from a model trained with single-stride gait features.

Unlike these studies, our work focuses on a multimodal solution that enables users to collect, create, and analyse a dataset of ground-level urban scenes. Our solution comprises a mobile application module and a computational library for loading and analysing the collected data. We believe this solution has the potential to facility research and data generation for urban experts.

\section{Methods}
\label{sec:methods}
This section presents the workflow of the methods adopted in this work (see Figure \ref{fig:workflow}), followed by a detailed description of the equipment and protocols used for dataset collection, as well as the types of data captured by the mobile application. 
The widespread availability of smartphones makes them ideal for transforming our solution into a user-friendly wearable tool. 

\input{figure/workflow}

Regarding the workflow, SideSeeing is a scientific initiative that involves the following steps: a) Collection and generation of multimodal datasets; b) Development of strategies for data preprocessing; c) Development of strategies for data visualization; d) Study, development and application of artificial intelligence models; and e) Development of strategies for information analysis.

\subsection{Equipment}
We opted for a low-cost chest mount (Figure \ref{fig:equipment-a}) to hold smartphones during data collection via the mobile application. This adjustable mount allows users to modify the angle between the device and their chest, facilitating a focus on either the ground or the horizon.

\input{figure/equipment}

Figure \ref{fig:equipment-b} depicts data collection with a 70-degree angle, ideal for capturing the ground near the user. In contrast, Figure \ref{fig:equipment-c} shows a 90-degree angle that focuses on the horizon and captures more urban furniture. This adjustable mount allows our solution to address both objectives.

For the SideSeeing dataset, we collected data using two different models of smartphones, both featuring accelerometer, gyroscope, and magnetometer sensors. For data collection, we recommend any smartphone with these sensors, and a rear camera with a minimum resolution of 1280x720 pixels capable of recording at 30 frames per second.

\subsection{Collection Protocol}
\label{section:collection-protocol}
For the data collection, we developed an Android-based application that utilizes Google's sensor and camera frameworks. The open-source application, named ``MultiSensor Data Collection'', is publicly available\footnote{\url{https://github.com/rafaelpezzuto/multi-sensor-data-collection}, accessed on June 10, 2024.}. It offers various settings that can be customized to a study's specific needs - Figure \ref{fig:mobile-app} presents screenshots of the mobile app. For the SideSeeing Hospital Dataset, the data collection protocol involved defining a set of parameters, including walking paths, starting angles configured in the mobile application, and device orientation. The following sections will explain the rationale behind our choices for building the dataset using this protocol.

\input{figure/mobile-app.tex}

Initially, we mapped out the routes in the four cities where the project participants are located (Santos, Jundiaí, São Paulo, and Chicago). As one of our project goals is to provide information regarding accessibility, we chose to collect data near hospitals, facilities where improper urban access can severely affect people's health. Through the Google Maps platform, we searched for bus stops and train or metro stations near the hospitals we had decided to cover. 
Each route starts at a public transportation stop and ends at the main entrance of a hospital (or vice-versa). The goal was to simulate a person's route traveling to and from the hospital. Figure \ref{fig:routes} illustrates two paths: one showing a path for a person walking from a hospital in Chicago, Illinois, USA to a public transportation stop (Figure \ref{fig:routes-a}), and the other showing a second path for a person walking from a public transportation stop to the hospital's entrance (Figure \ref{fig:routes-b}).

\input{figure/routes}

To facilitate adjusting the recording angle, the mobile application presents the angle value to the user before starting the data acquisition. The objective is to maximize the portion of the video showing the sidewalk. This configuration relies on using the default back camera of the device, which must have a wide enough field of view to capture the entire sidewalk width. We aim to utilize only the available cameras on mobile devices, avoiding the need for additional hardware or specific mobile devices (such as fisheye cameras). After a set of initial trials, we opted for an angle of approximately 70 degrees to focus on the ground, ensuring optimal sidewalk capture during the recordings.

Another parameter is the device orientation, which can be landscape or portrait. We tested both options, but in the context of the SideSeeing Hospital Dataset, we opted to use the landscape mode. We argue that by using the mobile application in this mode and with an adequate application angle, the video shows a greater level of detail and proximity to where the user is walking. Compounding with the other types of information collected by the mobile application, the potential to enrich the analysis regarding the sidewalk tends to be greater, enabling a more detailed examination of the sidewalk's surface material and conditions.

\begin{figure}[htp]
  \centering
  \includegraphics[width=1.\linewidth]{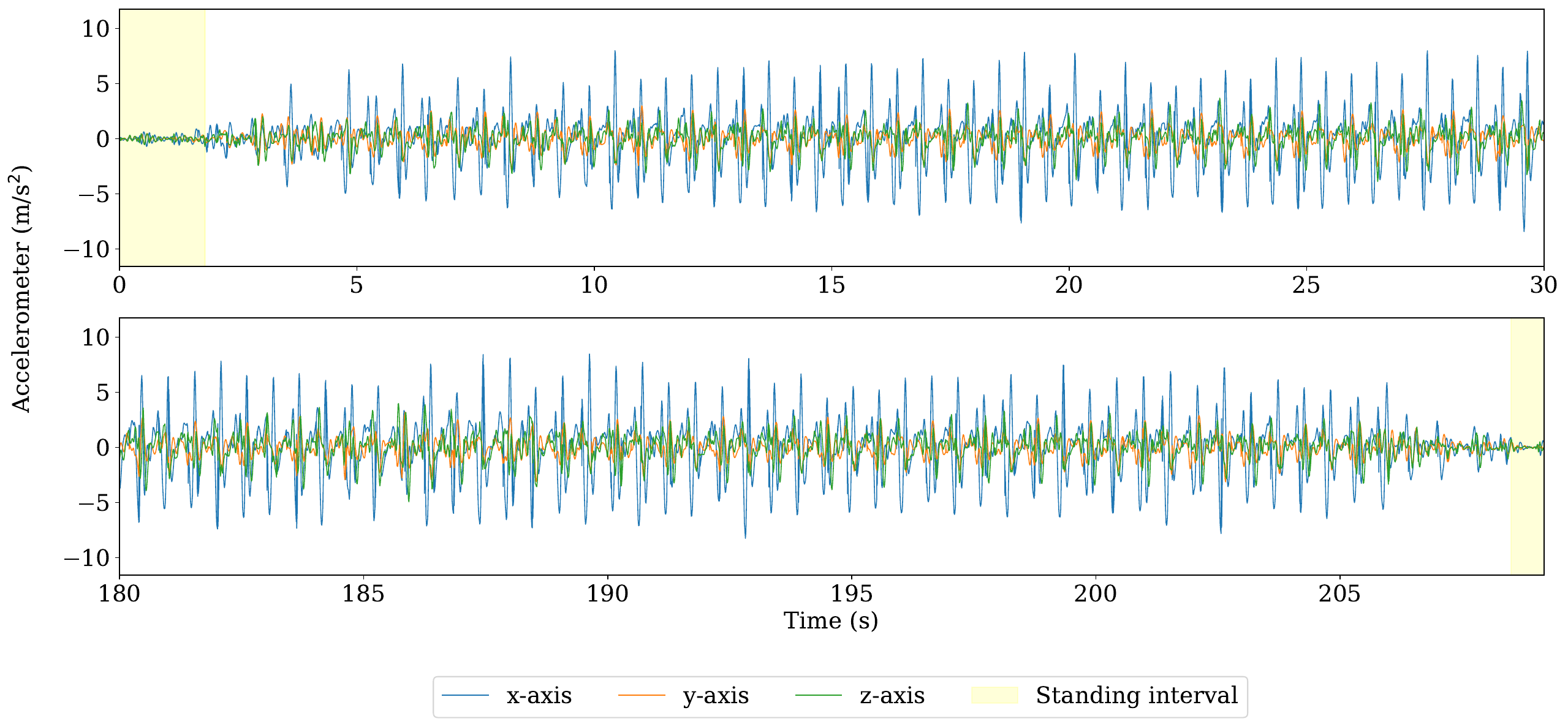}
  \caption{Accelerometer data revealing pause (yellow regions) and walking periods during a walking path. These data points were extracted from a route recorded in Chicago, Illinois, USA.}
  \label{fig:procedure-pauses}
\end{figure}

For data acquisition, we wore the chest support, configured the device by activating its auto-rotation and localization options, opened the mobile application, and started recording. The procedure involves pausing for two seconds (approximately), walking along the planned route, and pausing for another two seconds. As shown in Figure \ref{fig:procedure-pauses}, these pauses are particularly useful for post-editing steps based on the accelerometer data, such as video splitting. These exact steps were conducted for all the routes.

\subsection{Data Specification}
\label{section:data-specification}
The developed mobile application is capable of collecting four types of data, as follows: video (images and audio), time-series data from the device's sensors, and geographic localization data (latitude and longitude). In the case of our dataset, we decided to record data at 30 frames per second for the video with a resolution of 1280x720 pixels or 1920x1080 pixels, 15 points per second for the geographic information, and 50 Hertz for the device's sensors.

All data is stored locally on the device and can then be copied to the user's computer. In future versions, the application will automatically copy the data to a private cloud infrastructure. The sensor data and geographic data are stored in tabular format. The video includes audio and is stored in the well-known media format MP4. Data is  synchronized in terms of the device's operational system time. Table \ref{tab:data-specs} presents an overview of the files created during a collection of an instance.

\input{table/data-specs}

The file ``consumption.csv'' shows the current device's battery state. ``gps.csv'' contains geolocation coordinates, useful for tracking where the data was recorded (and even synchronizing it with information from other sources). Metadata, stored in a file named ``metadata.json'', keeps track of the device and application settings, including the recording start angle and user start and stop times, sensor frequency, camera resolution, and more. 

Sensor data files are named according to the number of axes they contain. For example, the single-axis proximity sensor data is stored in ``sensors.one.csv''. Three-axis sensors such as accelerometers, gyroscopes and magnetometer use ``sensors.three.csv''. For more information on sensors, refer to the official Google documentation\footnote{\url{https://developer.android.com/develop/sensors-and-location/sensors/sensors_overview}, accessed on June 10, 2024.}.

To further illustrate sensor information from Table \ref{tab:data-specs}, we present a plot of the gyroscope data in \ref{fig:gyroscope-a}. The yellow highlighted areas on the time series indicate a left turn made by the person recording the data. Similarly to the accelerometer case discussed earlier, this information is valuable for users to divide videos based on pedestrian movements. Figure \ref{fig:gyroscope-b} shows video frames corresponding to those moments when the pedestrian made turns.

\input{figure/gyroscope}

%

\section{SideSeeing Hospital Dataset}
\label{sec:dataset}
This section presents an overview of the SideSeeing Hospital Dataset, which focuses on sidewalks near public hospitals located in three Brazilian cities (e.g., Jundiaí, São Paulo, and Santos in the state of São Paulo) and one city in the USA (Chicago, Illinois). The goal was to collect data representing the possible routes from public transportation stops or stations to hospitals that elderly people and individuals with disabilities might need to take.

The dataset is composed of 46 videos, covering a distance of 12 kilometers, representing data collected in the four cities during January and June 2024. Table \ref{tab:dataset} presents detailed information regarding the data collected, grouped by city.

\input{table/dataset}

One possible use of our dataset is to enable studies that characterize sidewalks based on their surface material types. For instance, by analyzing video frames (as shown in Figure \ref{fig:surface-material-types}), researchers can observe the different materials used for sidewalk construction. Beyond manual observation, it's also possible to build automatic classifiers leveraging existent computer vision models (work in progress on our end).

\input{figure/surface-material-types}

The first row contains sidewalks with a cement surface.  From the beginning of row two until the second column of row three, the sidewalks have tactile pavement.  Row three contains sidewalks constructed with tiles. The last row contains a few more pavements, and we can note the sidewalk based on pavement stones, specifically Portuguese stones. 


Following the SideSeeing Hospital Dataset collection, we perform a manual analysis of the recorded videos to identify the various elements that compose the urban scene. Informed by this analysis and drawing on recent studies~\citep{gamache2017taxonomy, duan2021sidewalk, li2021taxonomy}, we constructed a two-level taxonomy focusing on sidewalk characteristics, specifically pertaining to pavement conditions, surface materials, obstacles, structures, and adjacent road types. A comprehensive list of these taxonomy elements is provided in the Appendix (Table \ref{tab:taxonomy}). We plan to describe in a new dedicated manuscript how we built and developed this taxonomy.

\section{SideSeeing Tools}
\label{sec:tools}
To support the analysis of the data collected through the MultiSensor Data Collection mobile application, we developed a Python library called ``sideseeing-tools''\footnote{For more information on ``sideseeing-tools'', see \url{https://pypi.org/project/sideseeing-tools/}.} which can be installed through ``pip''  with the following command: ``pip install sideseeing-tools''. It's a collection of scripts designed to load, preprocess, and visualize data, handling all content types within each dataset instance.

From the data collected through the Android application, users organize the information into a set of instances, each compose of sensor, GPS, video, and audio data. All ``sideseeing-tools'' functions were tailored to handle the synchronization of this information using timestamps. The library includes methods to extract information from a single instance or an entire dataset, offering a summary, such as the one shown in Table \ref{tab:dataset}. 

This tool offers the following functions for data visualization: geolocation (plot\_instance\_map, plot\_dataset\_cities, and plot\_dataset\_map), audio (plot\_instance\_audio), video frames (plot\_instance\_video\_frames\_at\_times, and plot\_instance\_video\_frames), sensors (generate\_video\_sensor3), and combined sensors with audio (plot\_instance\_sensors3\_and\_audio). It is also possible to extract data snippets using a function called extract\_snippet. This function separates the data into specified period interval and automatically handles video and sensor splitting.

\section{Conclusion}
\label{sec:conclusions}
In this paper, we presented the first step towards an initiative for street-level data acquisition, loading, and analysis. We introduced the SideSeeing Hospital Dataset, a resource for researchers studying sidewalk accessibility near hospitals. 
This dataset offers synchronized video footage, sensor data (accelerometer, gyroscope, magnetometer, GPS), and geographic location information. The data covers various sidewalk features, including pavement types, surface conditions, and surrounding elements.
The project\footnote{For further details regarding SideSeeing, please visit our website at \url{https://sites.usp.br/sideseeing/}.} also provides a publicly available framework for data acquisition (``MultiSensor Data Collection application'') and a Python library (``sideseeing-tools'') for data analysis. With these tools and dataset, researchers can conduct in-depth evaluations of sidewalk accessibility in diverse urban environments or create their own datasets.

As future work, we intend to capture data for other cities in North and South America.
Moreover, we plan to investigate the use of our multimodal data for gait analysis~\citep{shahar2021gaitanalysis, wang2022falldetection}, and classifiers leveraging existing computer vision models.
Given the complexity of SideSeeing's data, we also plan to explore human-centered approaches for labeling and evaluating the acquired data.

\section*{Acknowledgment}
The authors are grateful to FAPESP grant \#2022/15304-4, CNPq, CAPES, FINEP, MCTI PPI-SOFTEX (TIC 13 DOU 01245.010222/2022-44), Discovery Partners Institute, NSF (\#2320261, \#2330565), and IDOT (TS-22-340).

\bibliographystyle{plainnat}
\bibliography{references}

\newpage
\appendix
\section{Taxonomy of the SideSeeing Hospital Dataset}
\label{app:taxonomy}
\input{table/taxonomy}

\end{document}

%% file: figure/workflow.tex
\begin{figure}[htp]
  \centering
  \includegraphics[width=1\linewidth]{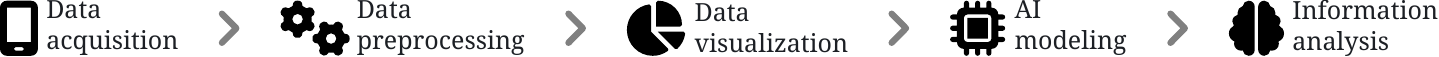}
  \caption{Workflow diagram showing the steps of SideSeeing from data acquisition to analysis.}
  \label{fig:workflow}
\end{figure}

%% file: figure/equipment.tex
\begin{figure}[htp]
  \centering
  \begin{subfigure}{.3\textwidth}
    \centering
    \includegraphics[width=\linewidth]{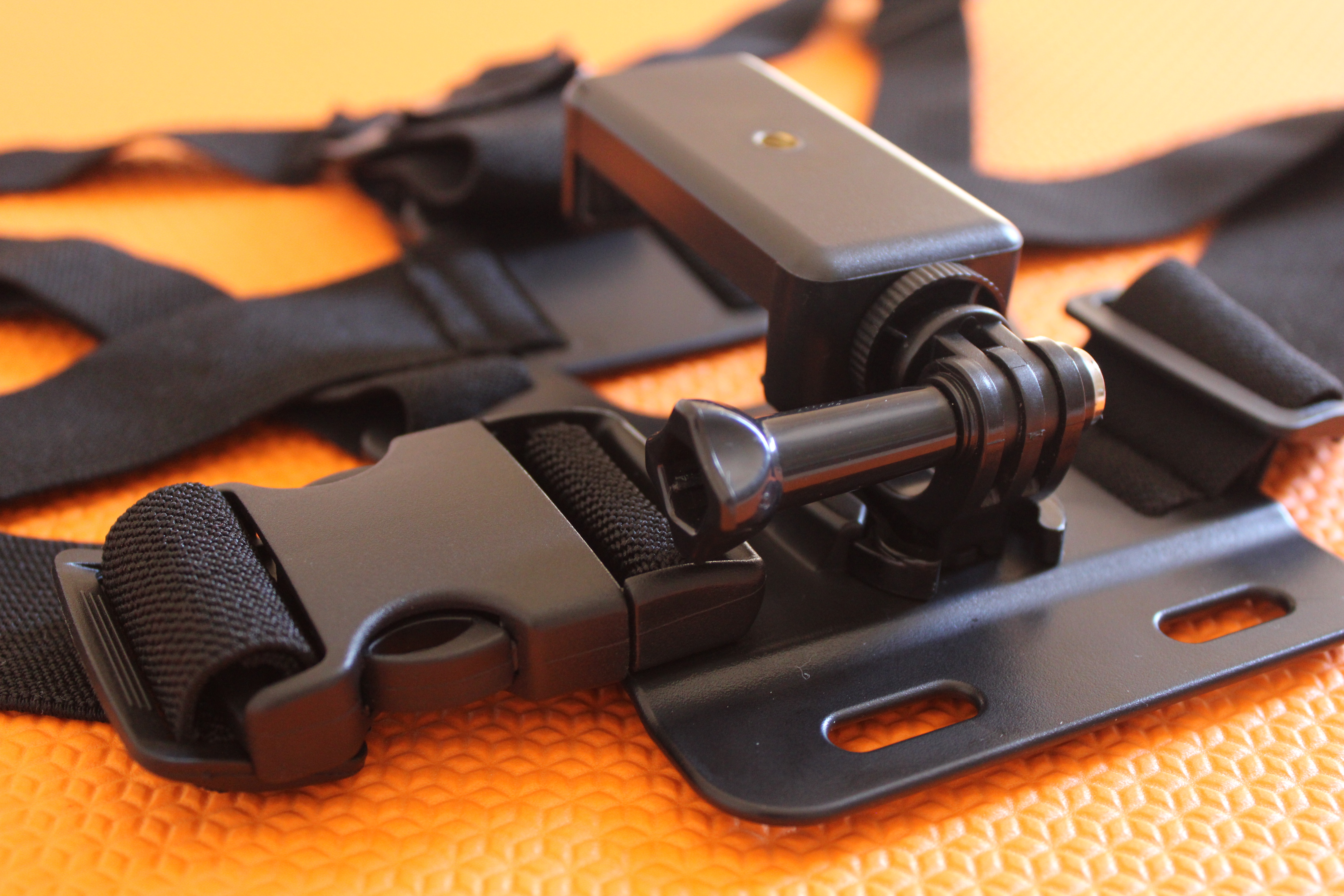}
    \caption{Chest mount for mobile device}
    \label{fig:equipment-a}
  \end{subfigure}
  \hfill
  \begin{subfigure}{.3\textwidth}
    \centering
    \includegraphics[width=\linewidth]{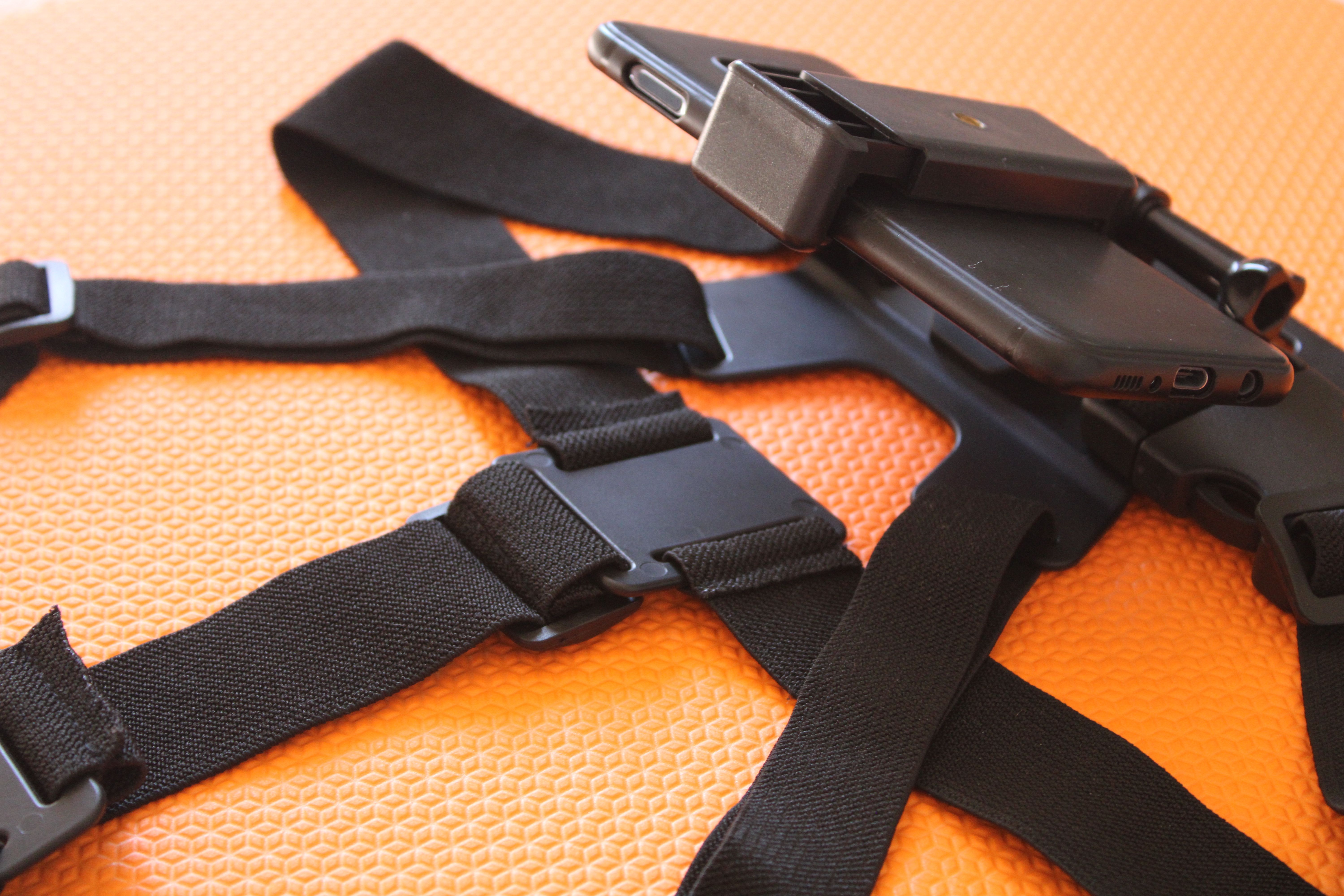}
    \caption{70-degree angle (ground view)}
    \label{fig:equipment-b}
  \end{subfigure}
  \hfill
  \begin{subfigure}{.3\textwidth}
    \centering
    \includegraphics[width=\linewidth]{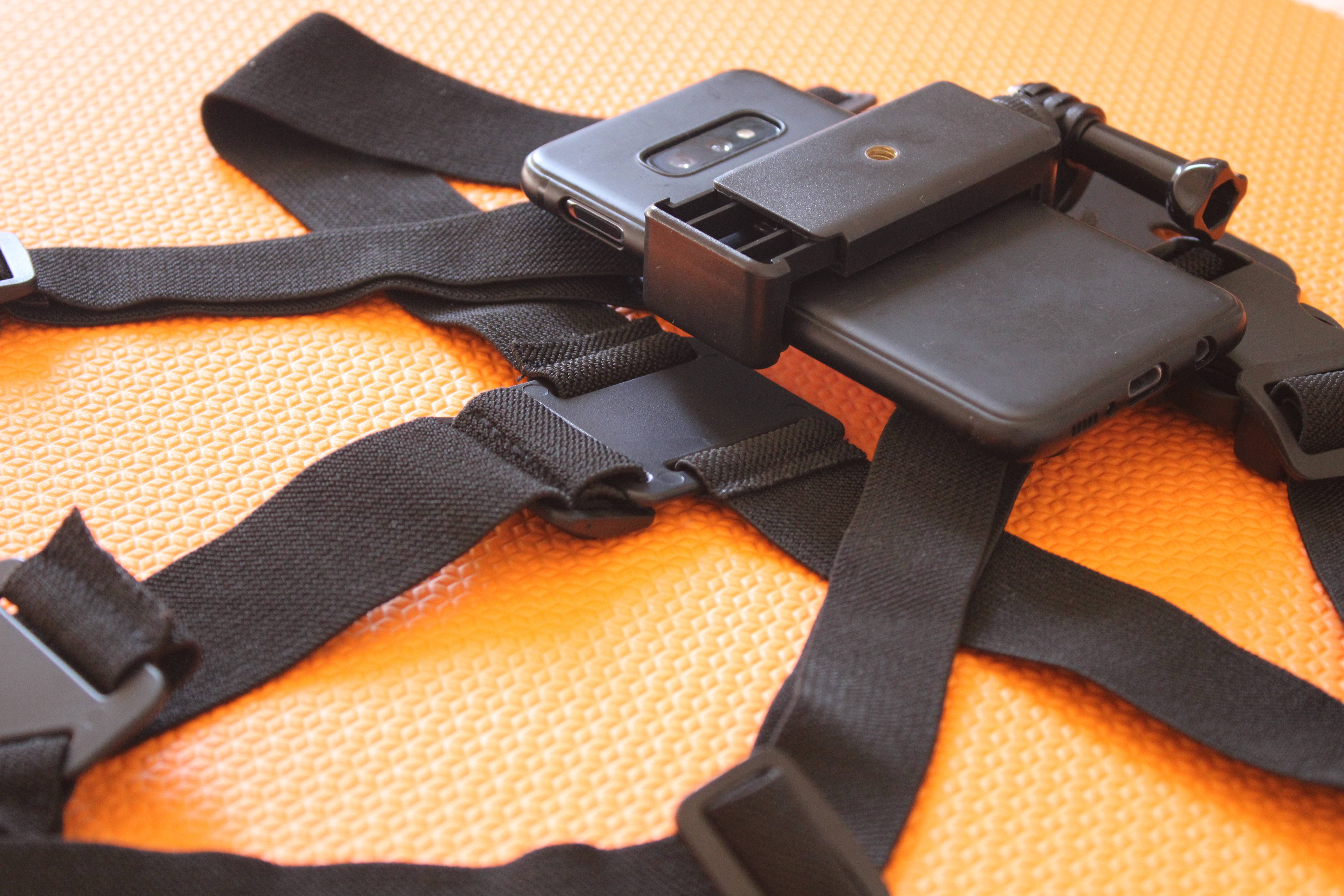}
    \caption{90-degree angle (horizon view)}
    \label{fig:equipment-c}
  \end{subfigure}
  \caption{Chest mount for mobile device with adjustable collection angle settings.}
\end{figure}

%% file: figure/mobile-app.tex
\begin{figure}[htp]
  \centering
  \begin{subfigure}{.49\textwidth}
    \centering
    \includegraphics[width=\linewidth]{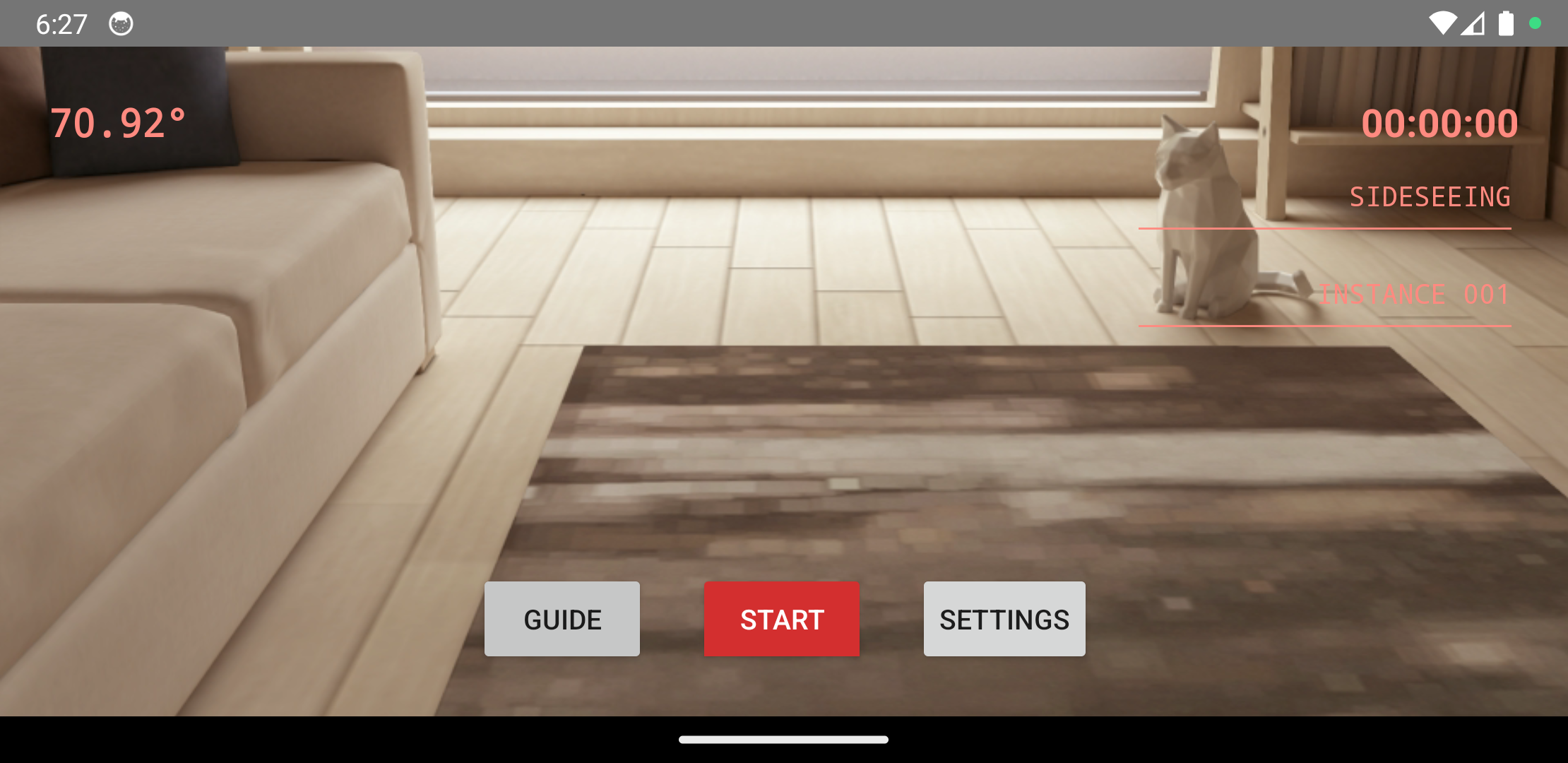}
    \caption{Main screen (pause state)}
    \includegraphics[width=\linewidth]{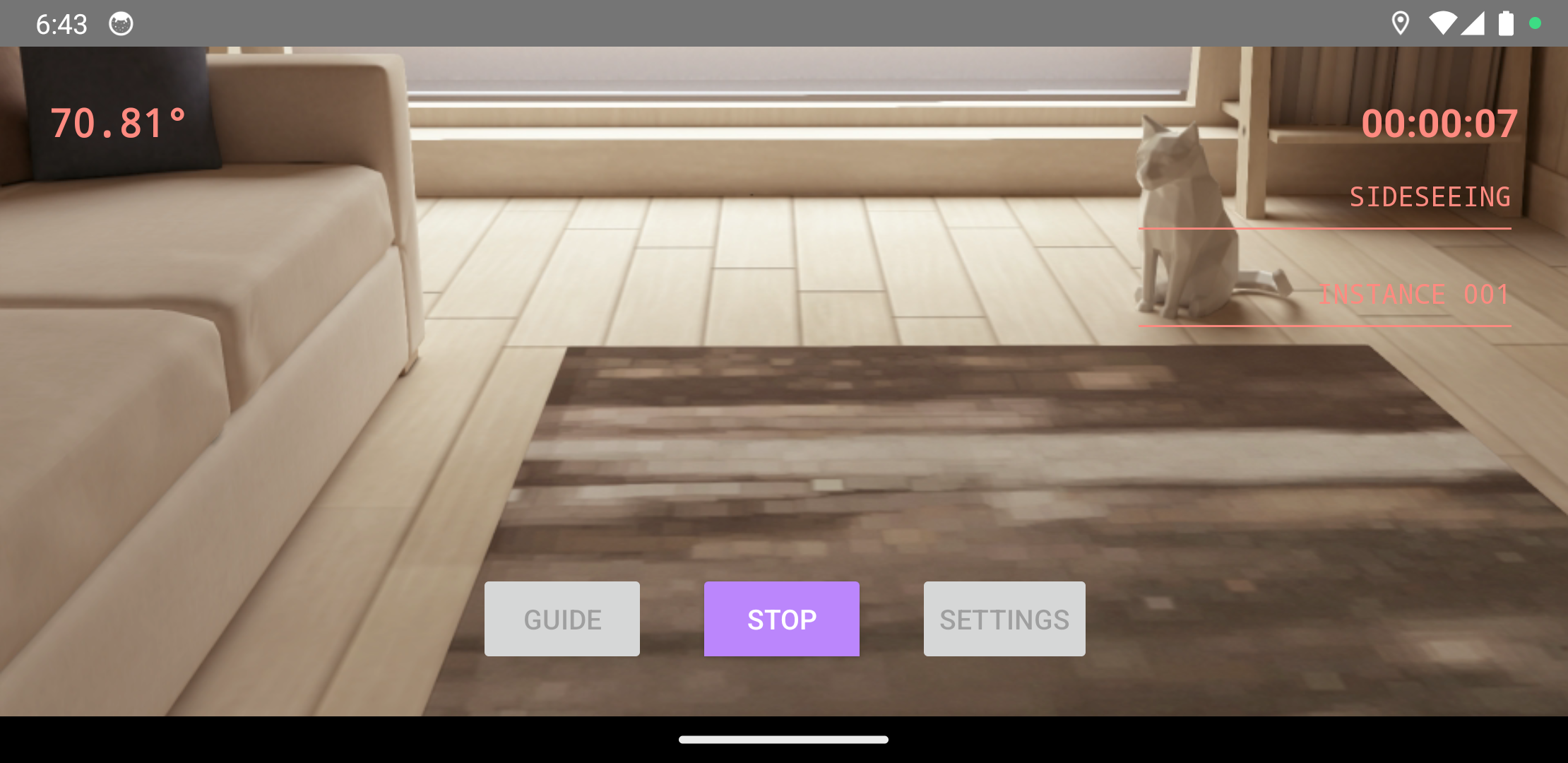}
    \caption{Main screen (recording state)}
  \end{subfigure}
  \begin{subfigure}{.2495\textwidth}
    \centering
    \includegraphics[width=\linewidth]{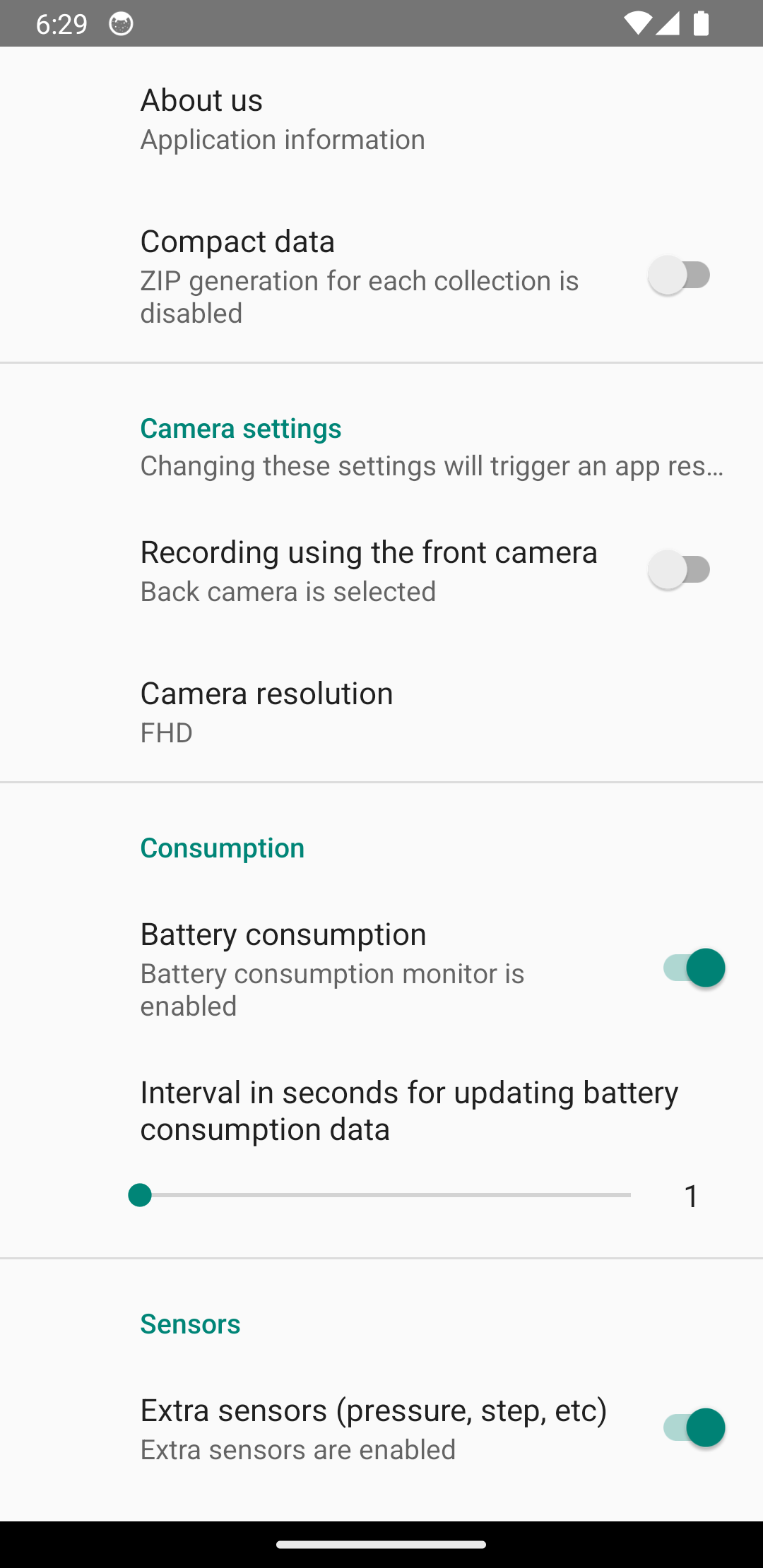}
    \caption{Settings screen, part one}
  \end{subfigure}
  \begin{subfigure}{.2495\textwidth}
    \centering
    \includegraphics[width=\linewidth]{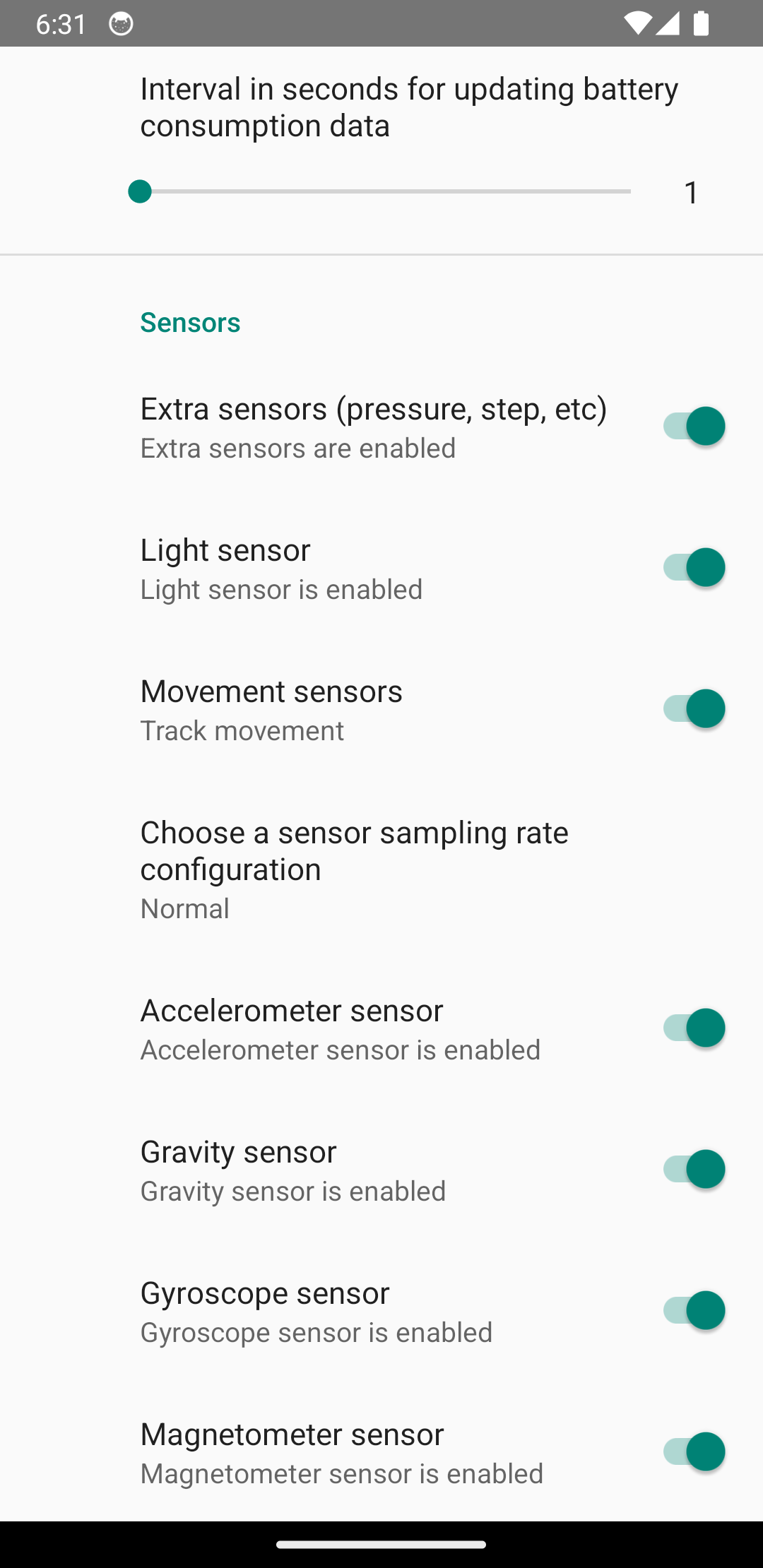}
    \caption{Settings screen, part two}
  \end{subfigure}
  \caption{Screenshots of the Android mobile application MultiSensor Data Collection.}
  \label{fig:mobile-app}
\end{figure}

%% file: figure/routes.tex
\begin{figure}[htp]
  \centering
  \begin{subfigure}{.45\textwidth}
    \centering
    \includegraphics[width=1.\linewidth]{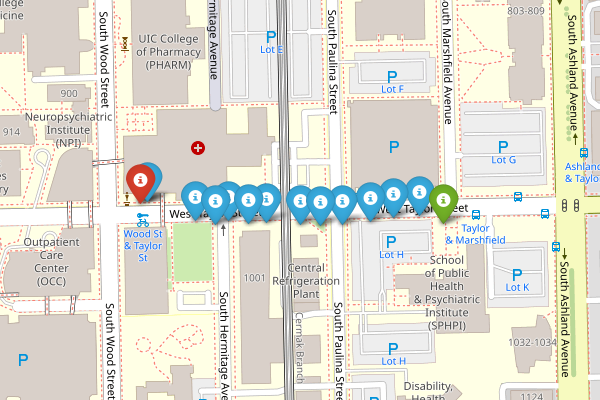}
    \caption{Route starting at a hospital.}
    \label{fig:routes-a}
  \end{subfigure}
  \hfill
  \begin{subfigure}{.45\textwidth}
    \centering
    \includegraphics[width=\linewidth]{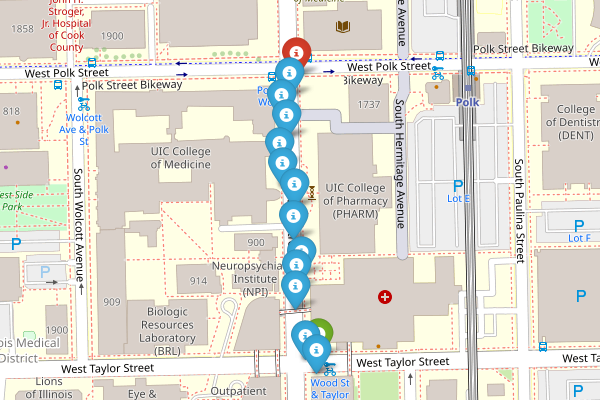}
    \caption{Route ending at a hospital.}
    \label{fig:routes-b}
  \end{subfigure}
  \caption{Example of routes traced for a person to walk between public transportation stops and a hospital in Chicago, Illinois, USA. The green marker represents the endpoint, and the red marker represents the starting point.}
  \label{fig:routes}
\end{figure}

%% file: table/data-specs.tex
\begin{table}[htp]
    \centering
    \caption{Files created by the MultiSensor Data Collection application for each collected instance.}
    \begin{tabular}{ll}
        \toprule
        File                            & Description \\
        \midrule
        consumption.csv                 & Data related to power consumption \\
        gps.csv                         & GPS data \\
        metadata.json                   & Metadata such as the device settings and application settings \\
        sensors.one.csv                 & Data from one-axis sensors \\
        sensors.three.csv               & Data from three-axis sensors \\
        sensors.three.uncalibrated.csv  & Data from three-axis sensors uncalibrated \\
        video.mp4                       & MP4 video file including audio \\
        \bottomrule
    \end{tabular}
    \label{tab:data-specs}
\end{table}

%% file: figure/gyroscope.tex
\begin{figure}[htp]
  \centering
  \begin{subfigure}{1\textwidth}
    \centering
    \includegraphics[width=1.\linewidth]{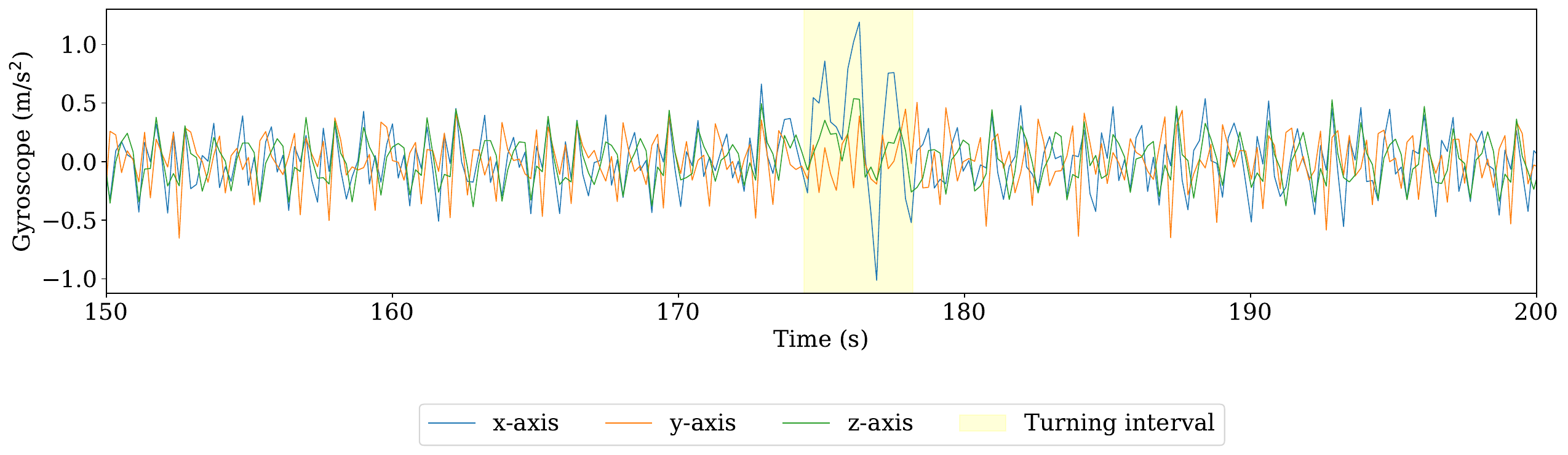}
    \caption{Gyroscope data highlighted in yellow for the moment when the pedestrian turns left.}
    \label{fig:gyroscope-a}
  \end{subfigure}
  \begin{subfigure}{1\textwidth}
    \centering
    \includegraphics[width=\linewidth]{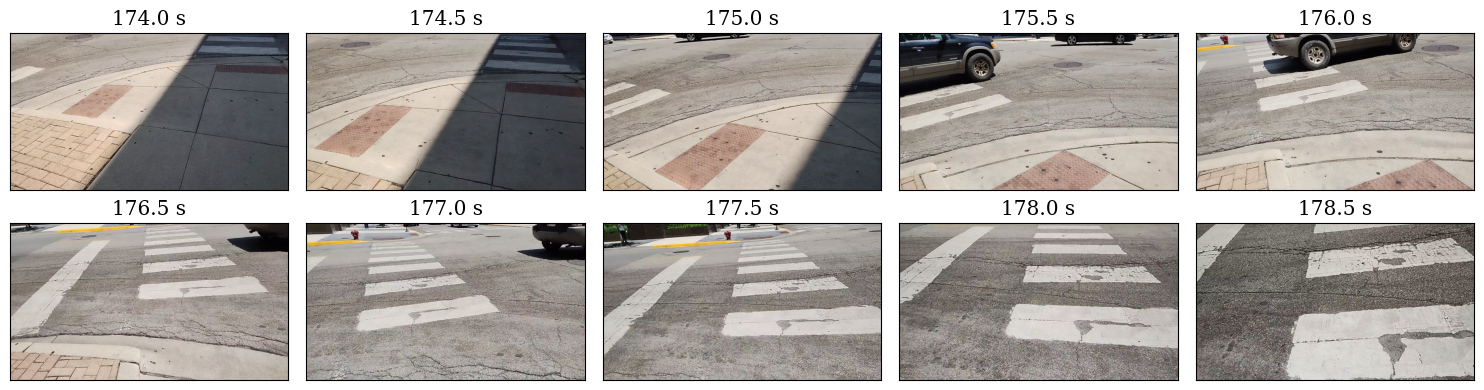}
    \caption{Video frames for the segment where the pedestrian turns left.}
    \label{fig:gyroscope-b}
  \end{subfigure}
  \caption{Gyroscope data revealing turns during a walking path. These data points were extracted from a route recorded in Chicago, Illinois, USA.}
  \label{fig:gyroscope}
\end{figure}

%% file: table/dataset.tex
\begin{table}[htp]
    \centering
    \caption{Total distance in meters, duration in seconds, number of routes, number of hospitals, total video frames, and total data points for Accelerometer (ACC), gyroscope (GYR), and magnetometer (MAG) across cities.}
\begin{tabular}{@{}rrrrrrrrr@{}}
\toprule
& \multicolumn{1}{r}{}  & \multicolumn{1}{r}{}  & \multicolumn{1}{r}{}   & \multicolumn{1}{r}{}   & \multicolumn{1}{r}{}   & \multicolumn{3}{c}{Total Data Points} \\ \cmidrule(l){7-9}
City - Country & \multicolumn{1}{r}{Routes} & \multicolumn{1}{r}{Hospitals} & \multicolumn{1}{r}{Distance} & \multicolumn{1}{r}{Duration} & \multicolumn{1}{r}{Video Frames} & \multicolumn{1}{r}{ACC} & \multicolumn{1}{r}{GYR} & \multicolumn{1}{r}{MAG} \\
\midrule
Chicago - USA      & 25 & 3 & 5,865  & 4,616  & 175,847 & 609,890   & 58,247  & 333,626 \\
Jundiaí - Brazil   & 7  & 2 & 1,482  & 1,890  & 44,476  & 139,294   & 8,243   & 68,096  \\
Santos - Brazil    & 11 & 2 & 2,247  & 3,532  & 67,431  & 215,573   & 12,499  & 105,680 \\
São Paulo - Brazil & 4  & 2 & 1,271  & 1,840  & 38,129  & 92,638    & 63,624  & 38,073  \\
\midrule
All                & 47 & 9 & 10,865 & 11,878 & 325,883 & 1,057,395 & 142,613 & 545,475 \\
\bottomrule
\end{tabular}
\label{tab:dataset}
\end{table}

%% file: figure/surface-material-types.tex
\begin{figure}[htp]
    \centering
    \subfloat{\includegraphics[width=0.25\textwidth]{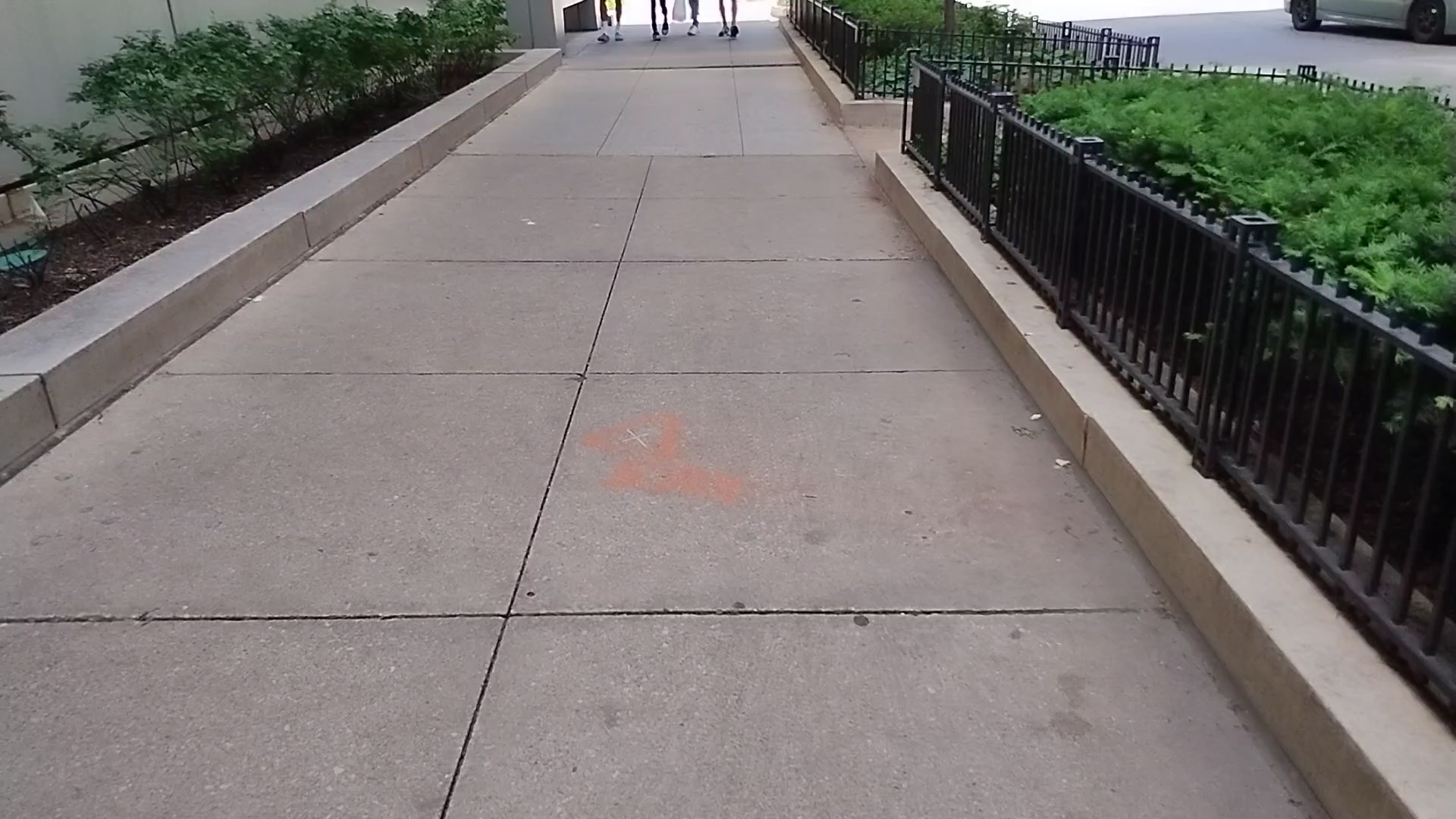}}
    \subfloat{\includegraphics[width=0.25\textwidth]{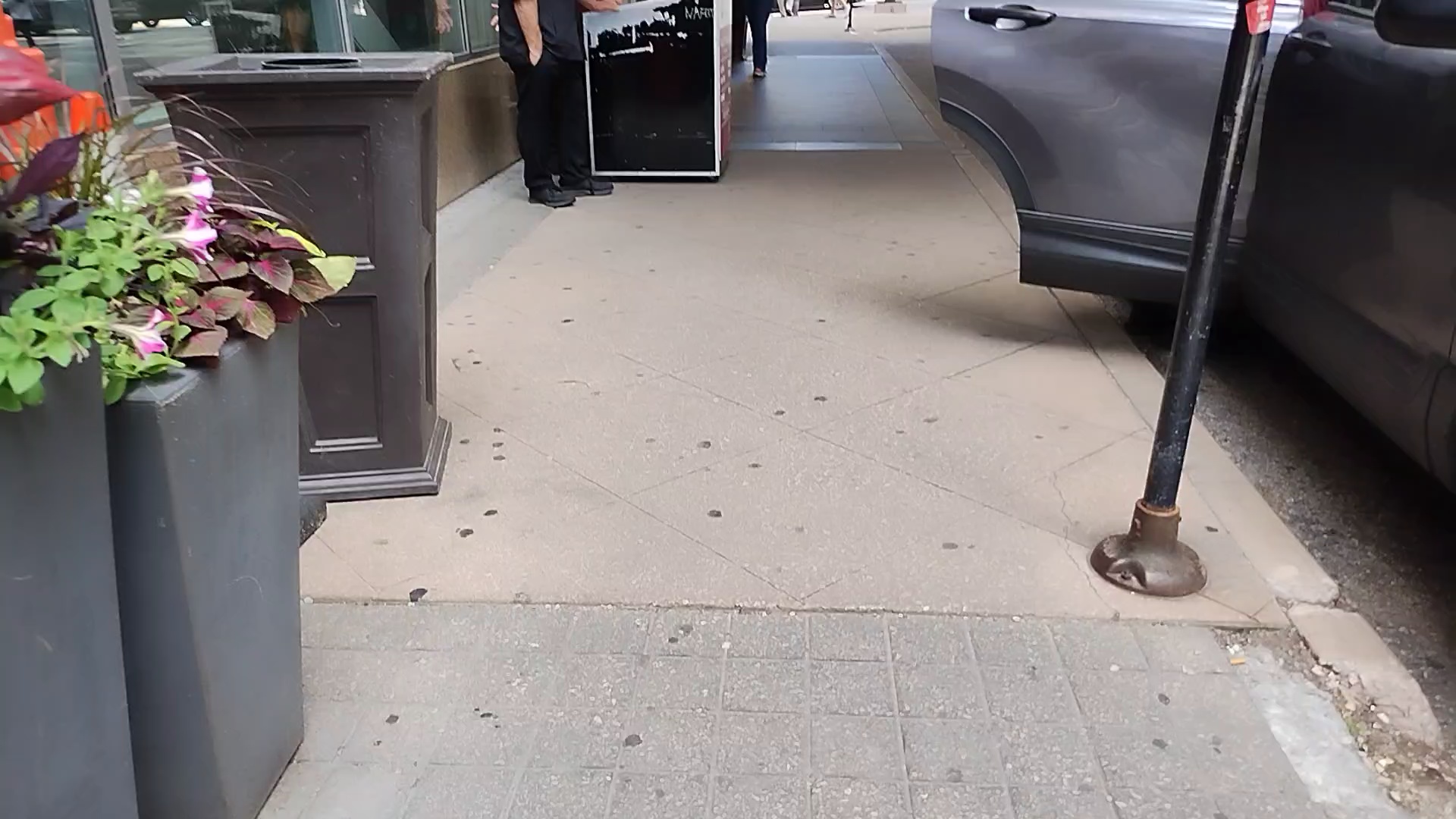}}
    \subfloat{\includegraphics[width=0.25\textwidth]{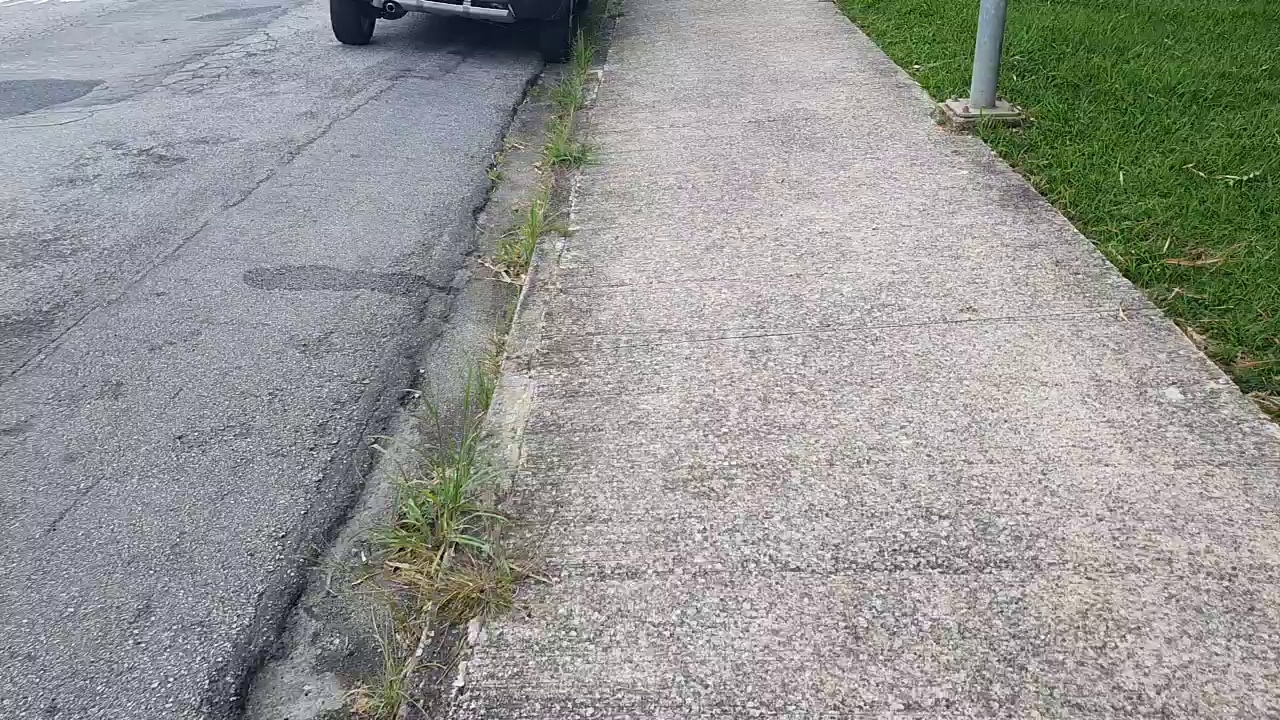}}
    \subfloat{\includegraphics[width=0.25\textwidth]{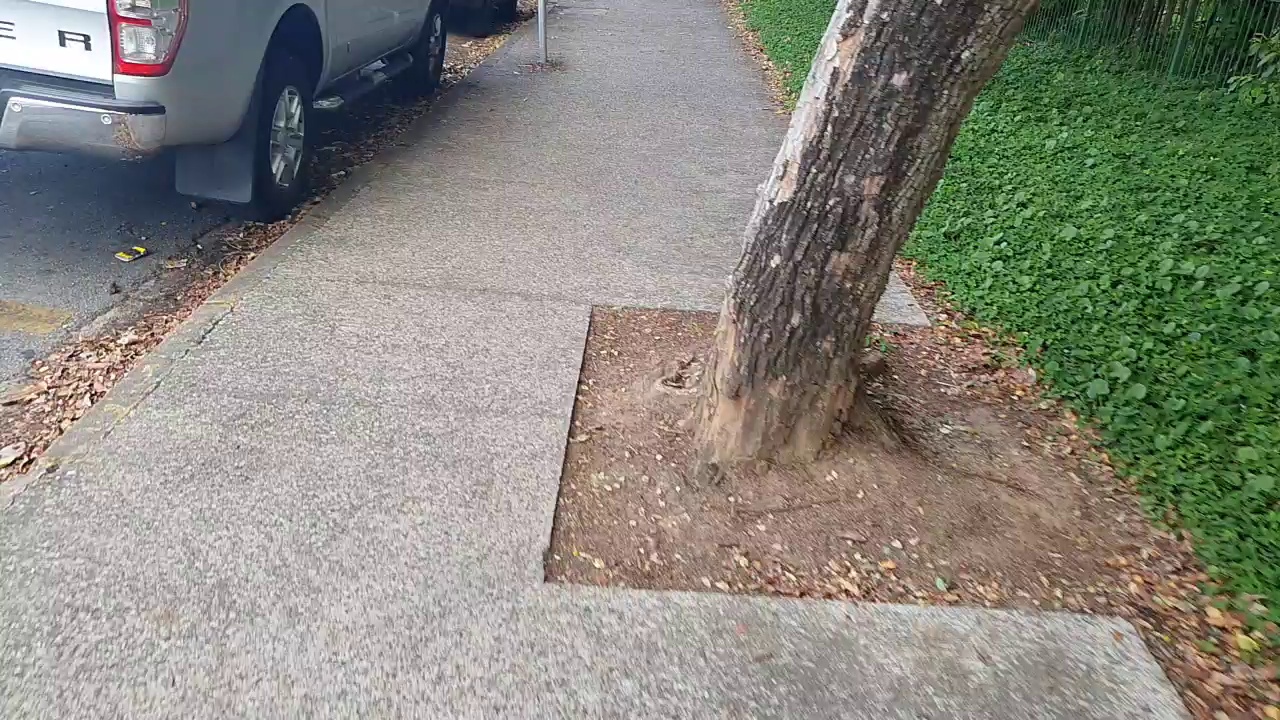}}
    \\\vspace{-0.1em}
    \subfloat{\includegraphics[width=0.25\textwidth]{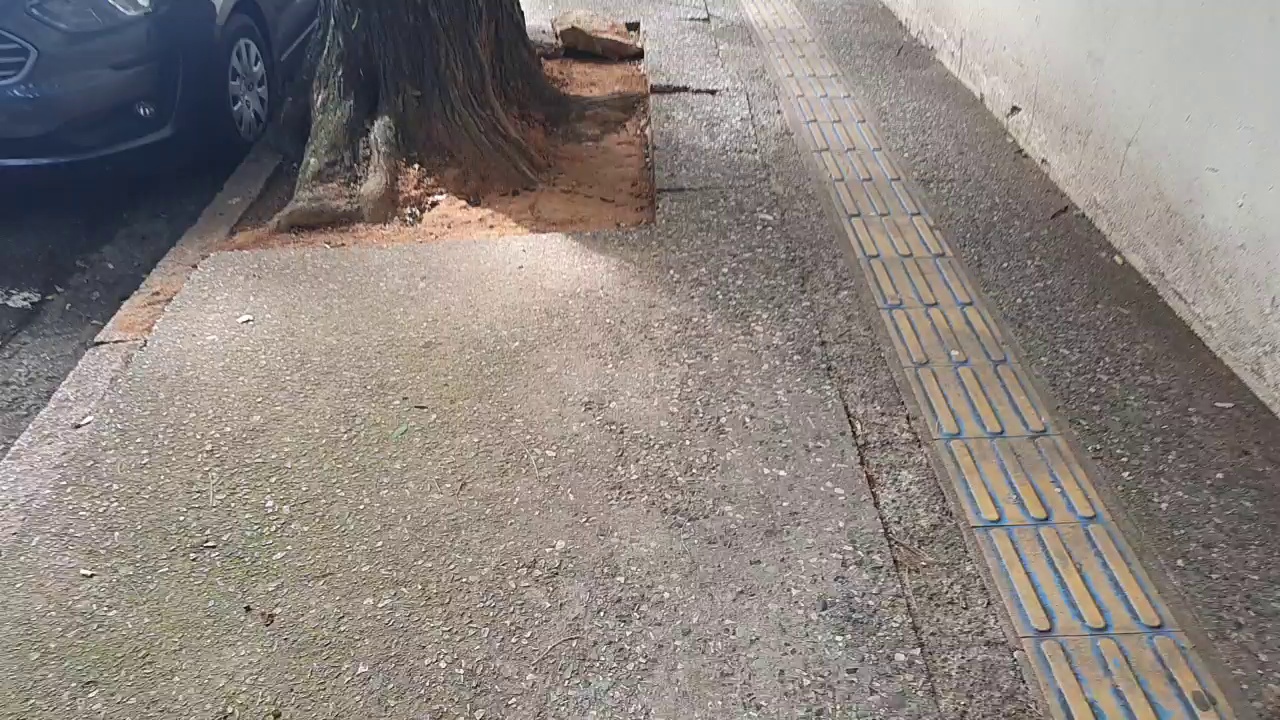}}
    \subfloat{\includegraphics[width=0.25\textwidth]{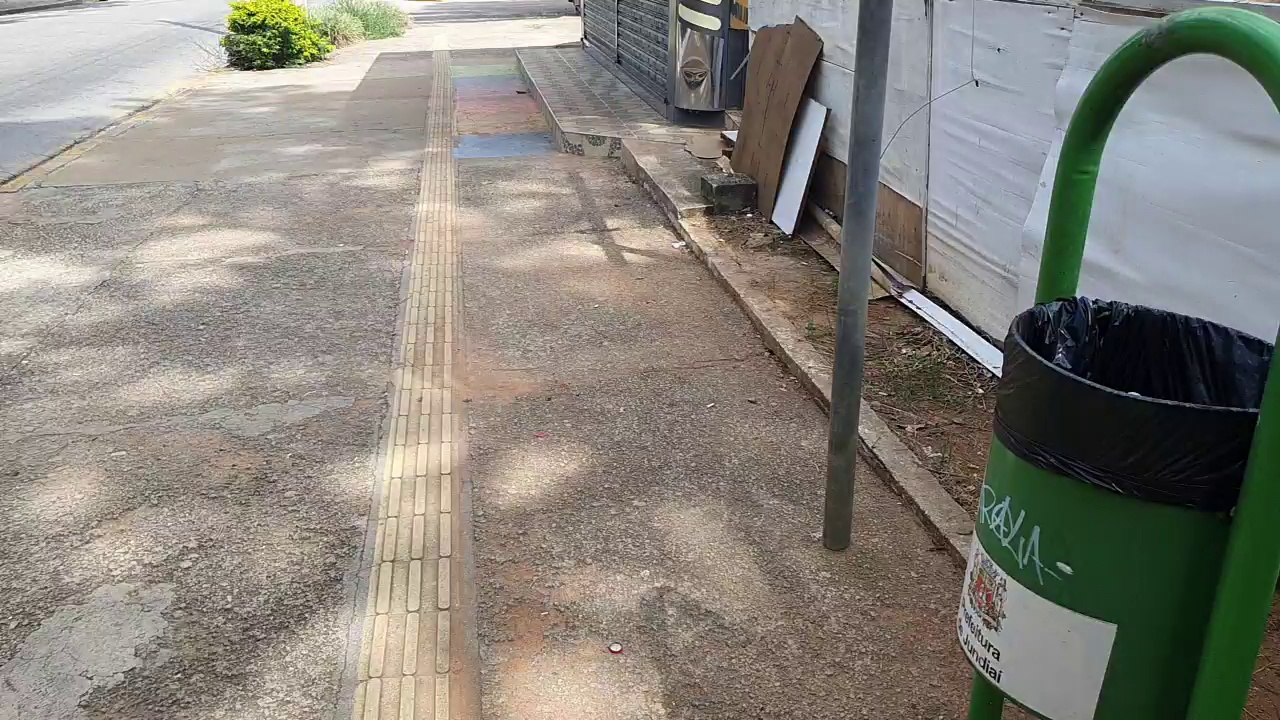}}
    \subfloat{\includegraphics[width=0.25\textwidth]{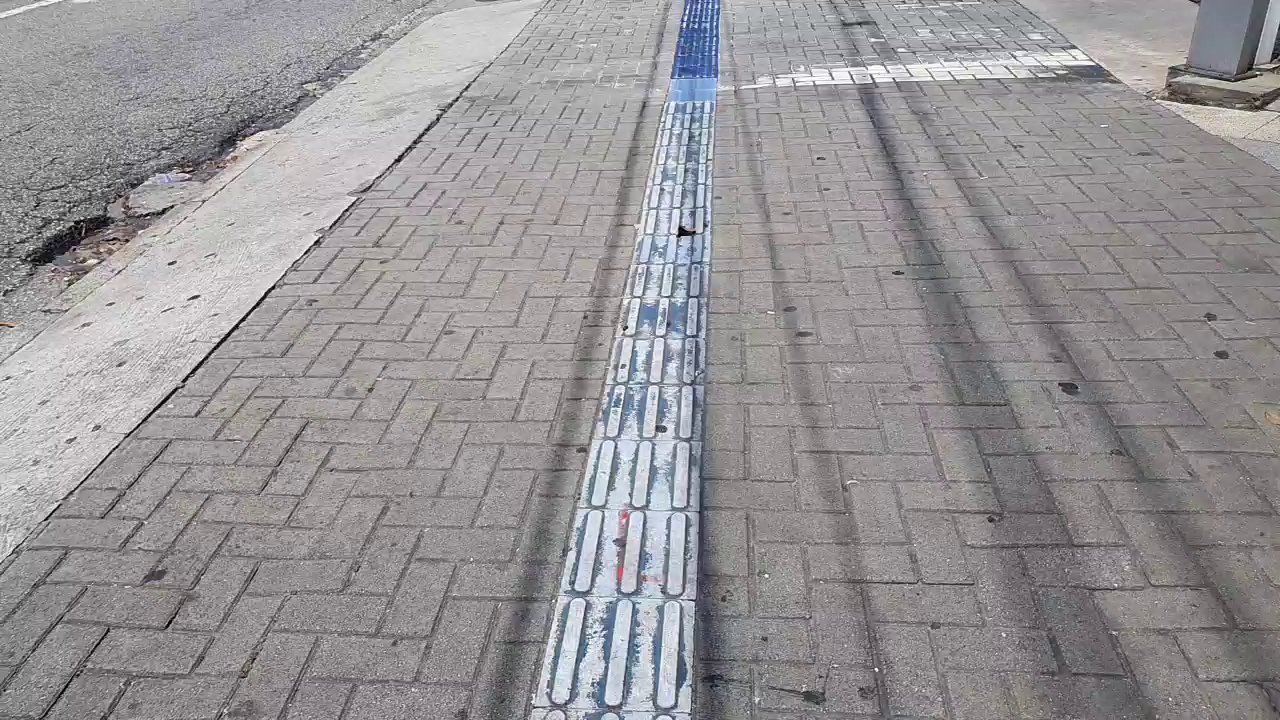}}
    \subfloat{\includegraphics[width=0.25\textwidth]{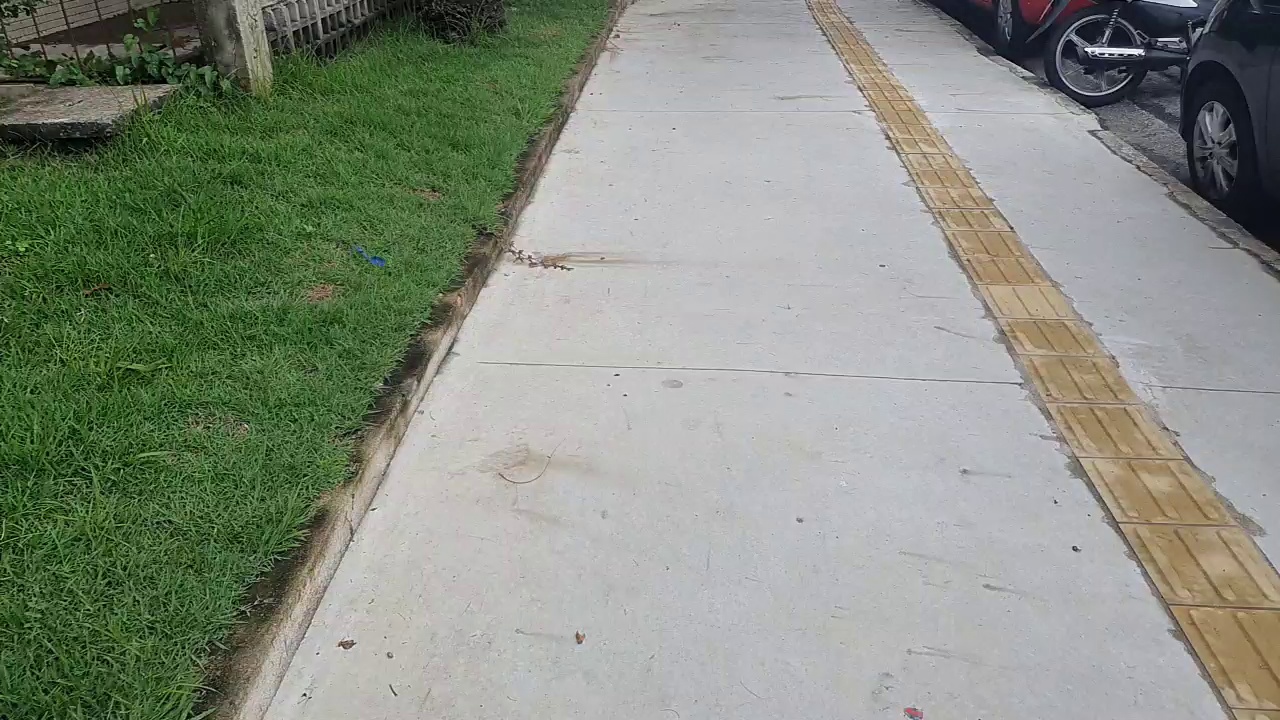}}
    \\\vspace{-0.1em}
    \subfloat{\includegraphics[width=0.25\textwidth]{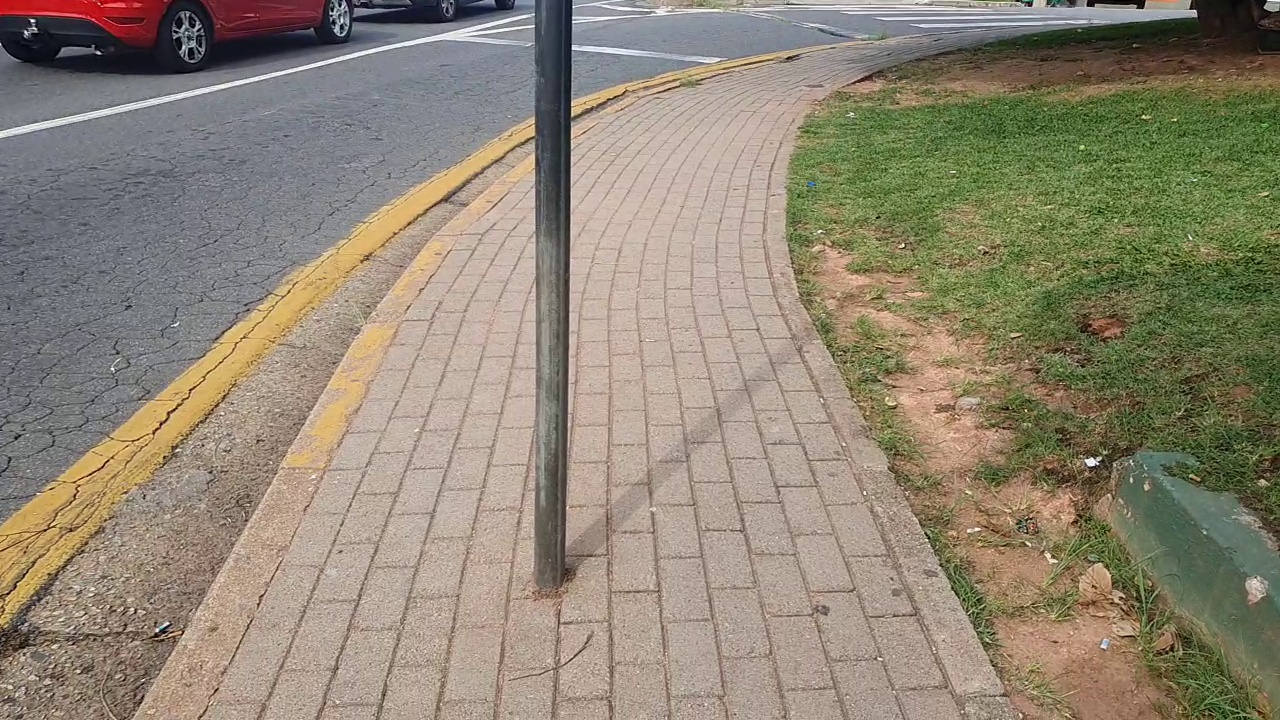}}
    \subfloat{\includegraphics[width=0.25\textwidth]{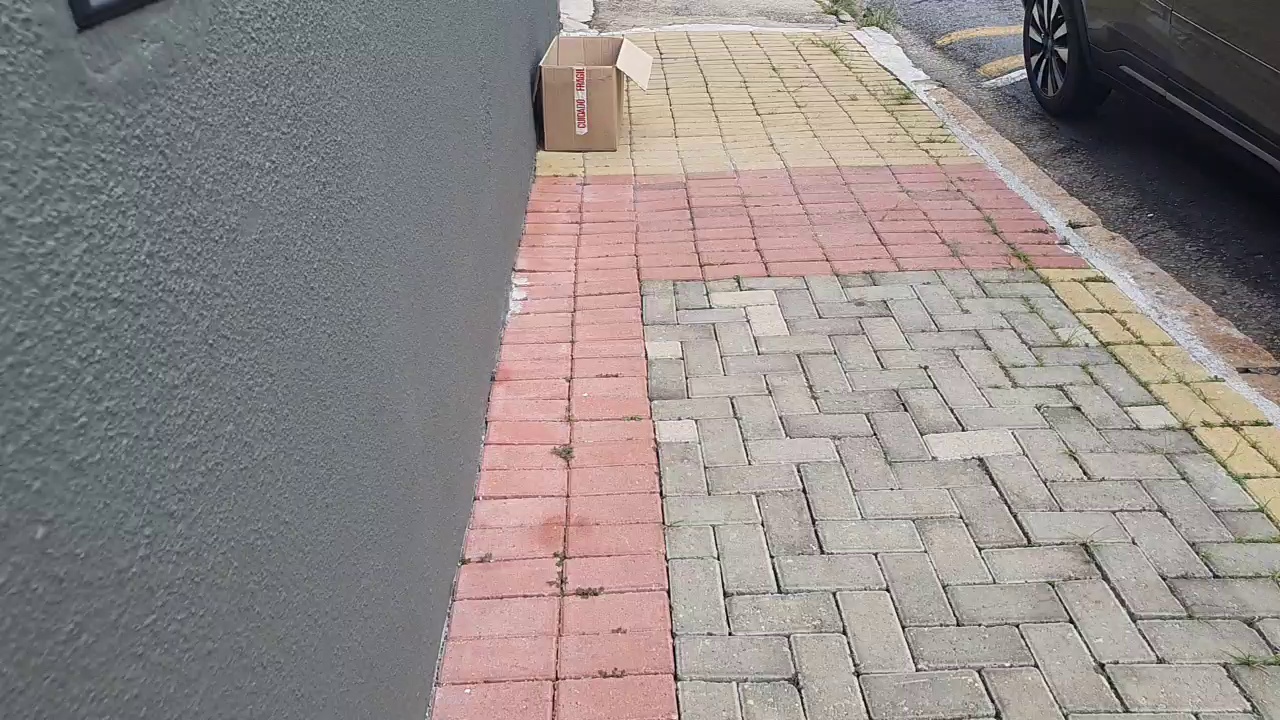}}
    \subfloat{\includegraphics[width=0.25\textwidth]{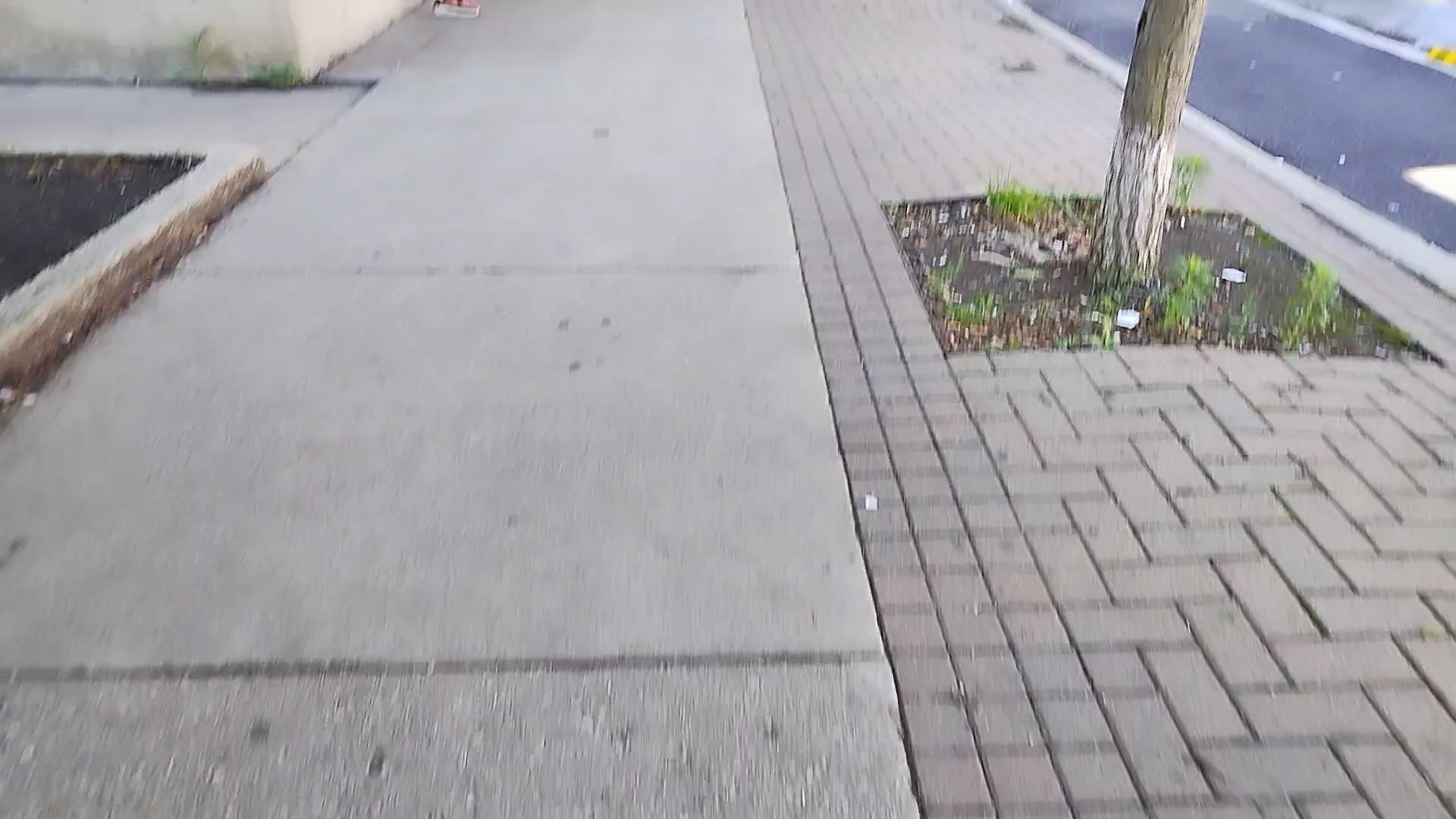}}
    \subfloat{\includegraphics[width=0.25\textwidth]{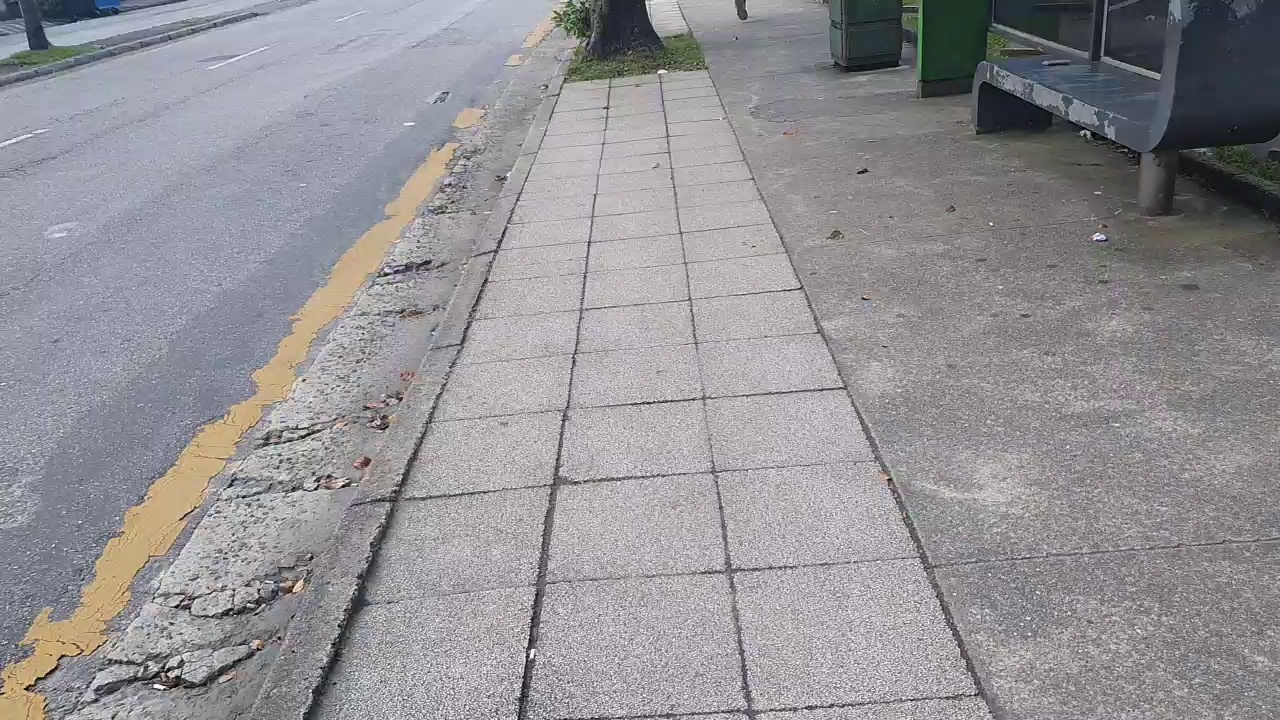}}
    \\\vspace{-0.1em}
    \subfloat{\includegraphics[width=0.25\textwidth]{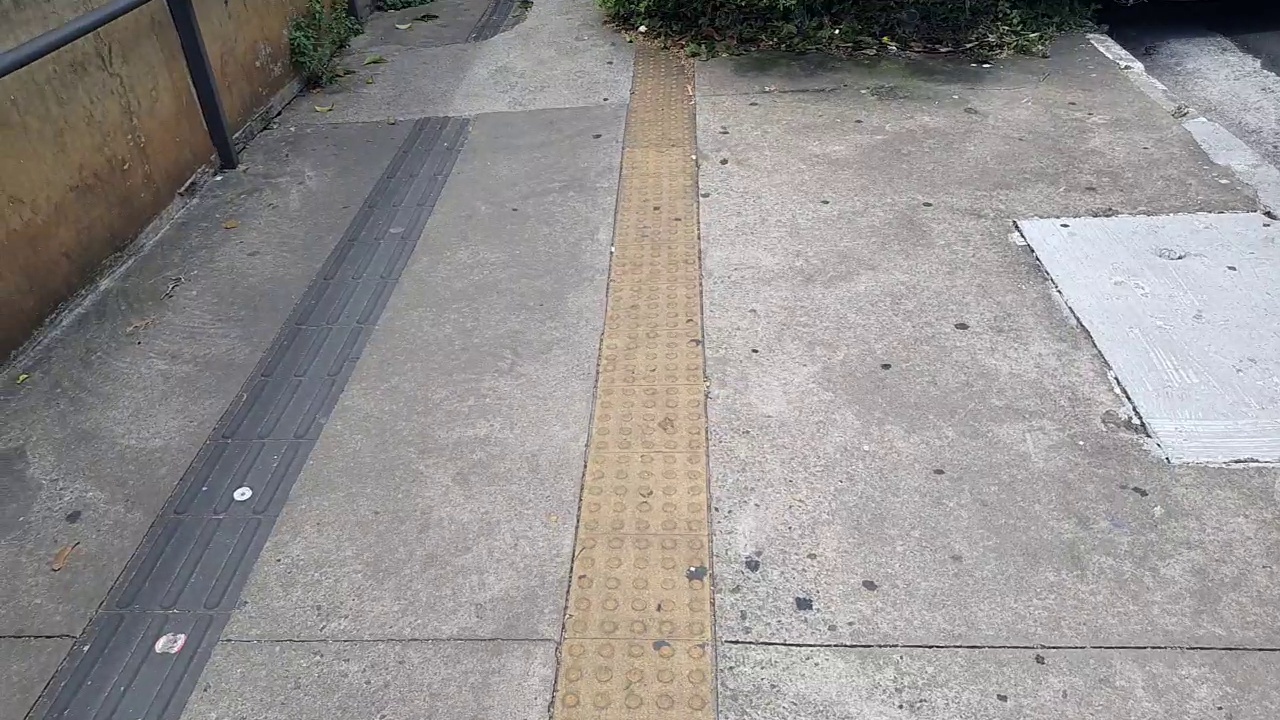}}
    \subfloat{\includegraphics[width=0.25\textwidth]{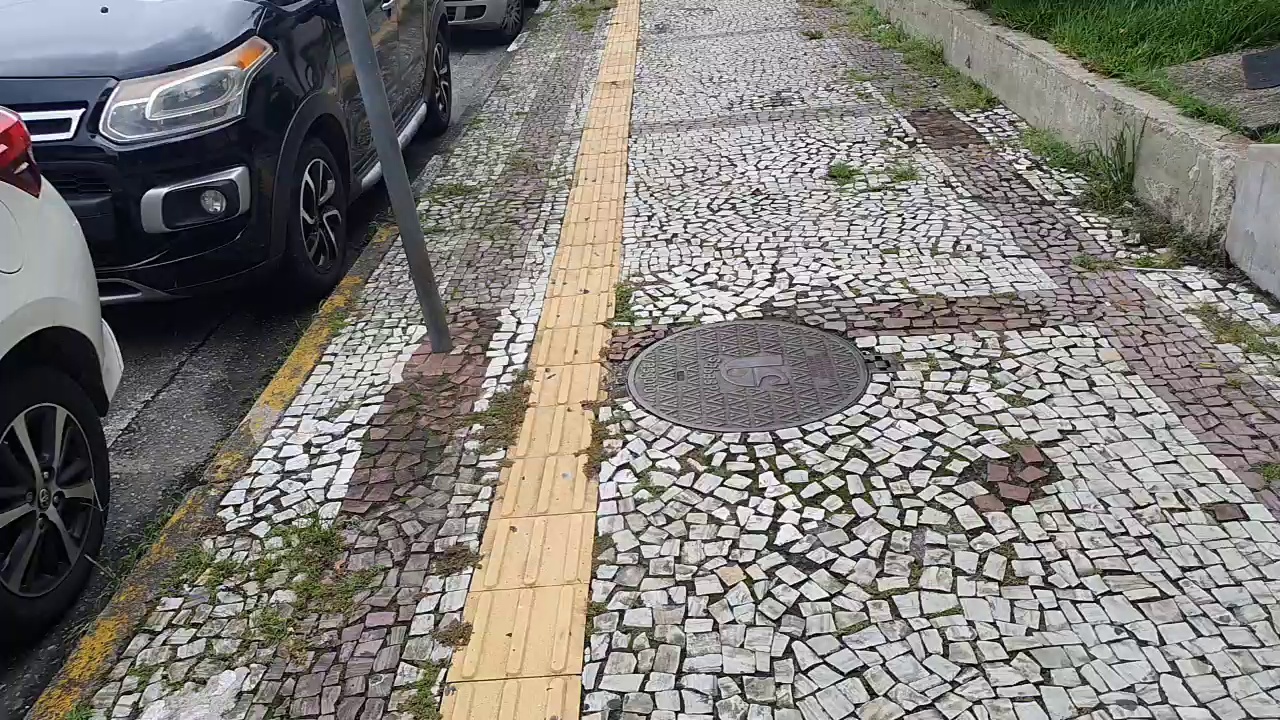}}
    \subfloat{\includegraphics[width=0.25\textwidth]{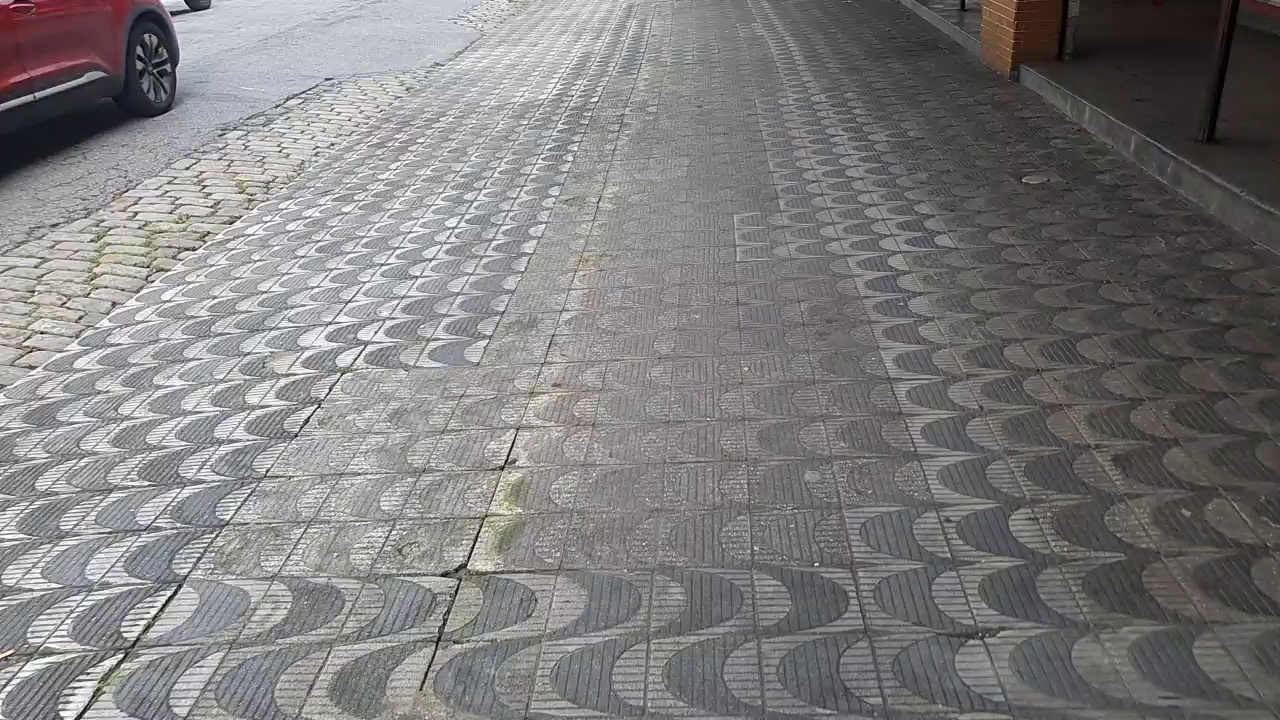}}
    \subfloat{\includegraphics[width=0.25\textwidth]{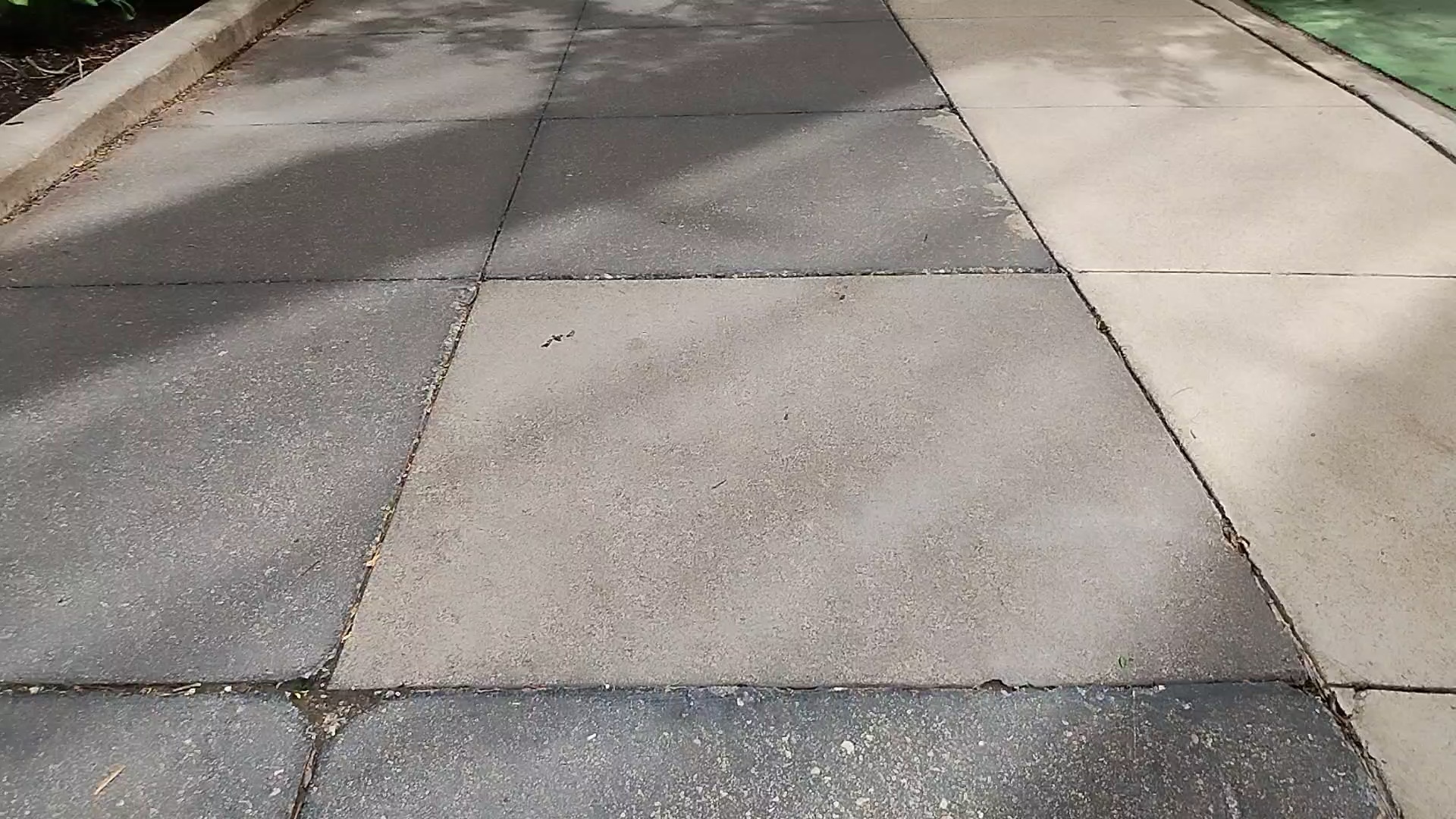}}
    \caption{Examples of sidewalk surface material types observed in our dataset.}
    \label{fig:surface-material-types}
\end{figure}

%% file: table/taxonomy.tex
\begin{table}[htp]
\centering
\caption{Taxonomy of sidewalks with two levels. Level one includes adjacent road types, obstacles, pavement conditions, sidewalk geometry, sidewalk structure, and surface materials.}
\begin{tabular}{ll|ll}
\toprule
Level 1             & Level 2              & Level 1            & Level 2                  \\
\midrule
Adjacent road type & Motorway / highway    & Pavement condition & Broken                   \\
                   & None                  &                    & Corrugation              \\
                   & Residential           &                    & Cracked                  \\
                   & Service               &                    & Detached                 \\
Obstacles          & Aerial vegetation     &                    & Patching                 \\
                   & Barrier               &                    & Pothole                  \\
                   & Bench                 & Sidewalk geometry  & Height difference        \\
                   & Bike rack             &                    & Narrow                   \\
                   & Black ice             &                    & Steep                    \\
                   & Bus stop              & Sidewalk structure & Bioswale                 \\
                   & Car barrier           &                    & Curb ramp                \\
                   & Construction material &                    & Footbridge               \\
                   & Dirt                  &                    & Friction strip           \\
                   & Fence                 &                    & Grate                    \\
                   & Fire hydrant          &                    & Ramp                     \\
                   & Floor standing board  &                    & Stairs                   \\
                   & Garage entrance       &                    & Tactile paving           \\
                   & Ground light          & Surface type       & Asphalt                  \\
                   & Ground vegetation     &                    & Bluestone                \\
                   & Manhole cover         &                    & Brick                    \\
                   & Newsstand             &                    & Coating                  \\
                   & Parked vehicle        &                    & Cobblestone              \\
                   & Parking booth         &                    & Concrete                 \\
                   & Person                &                    & Concrete with aggregates \\
                   & Pole                  &                    & Grass                    \\
                   & Potted plant          &                    & Gravel                   \\
                   & Puddle                &                    & Large pavers             \\
                   & Rock                  &                    & Red brick                \\
                   & Snow                  &                    & Slab                     \\
                   & Telephone booth       &                    & Stone pavement           \\
                   & Traffic cone          &                    & Tiles                    \\
                   & Transit sign          &                    & \\
                   & Trash can             &                    & \\
                   & Tree leaves           &                    & \\
                   & Trunck                &                    & \\
                   & Water channel         &                    & \\
                   & Water fountain        &                    & \\
\bottomrule                   
\end{tabular}
\label{tab:taxonomy}
\end{table}